\documentclass[]{fairmeta}

\usepackage{comment}
\usepackage{xspace}
\usepackage{marginnote}
\usepackage{xcolor} % Required for colored text
\usepackage[dvipsnames]{xcolor}
\usepackage{multirow}
\usepackage{nicematrix}
\usepackage{wrapfig}
\usepackage{amsmath}
\usepackage{amssymb}
\usepackage{amsfonts}
\usepackage{algorithm}
\usepackage{algorithmicx}
\usepackage{algpseudocode}
\usepackage{float}

\usepackage{nicematrix}  % Required for NiceTabular
\usepackage{graphicx}
\usepackage{booktabs}
\usepackage{enumitem}

\def\method{SAM 3D\xspace}
\definecolor{readableyellow}{RGB}{218, 165, 32}  % goldenrod

\makeatletter
\DeclareRobustCommand\onedot{\futurelet\@let@token\@onedot}
\def\@onedot{\ifx\@let@token.\else.\null\fi\xspace}
\def\eg{\emph{e.g}\onedot} 
\def\ie{\emph{i.e}\onedot} 

\renewcommand{\paragraph}[1]{\vspace{.5em}\noindent\textbf{#1}}
\title{SAM 3D: 3Dfy Anything in Images}

\author{SAM 3D Team}
\author[*]{Xingyu Chen}
\author[*]{Fu-Jen Chu}
\author[*]{Pierre Gleize}
\author[*]{Kevin J Liang}
\author[*]{Alexander Sax}
\author[*]{Hao Tang}
\author[*]{Weiyao Wang}
\author{Michelle Guo}
\author{Thibaut Hardin}
\author[\circ]{Xiang Li}
\author{Aohan Lin}
\author{Jiawei Liu}
\author[\circ]{Ziqi Ma}
\author{Anushka Sagar}
\author[\circ]{Bowen Song}
\author{Xiaodong Wang}
\author[\circ]{Jianing Yang}
\author[\circ]{Bowen Zhang}
\author[\dagger]{Piotr Doll\'ar}
\author[\dagger]{Georgia Gkioxari}
\author[\dagger\S]{Matt Feiszli}
\author[\dagger\S]{Jitendra Malik}

\affiliation{Meta Superintelligence Labs}

\contribution[*]{Core Contributor (Alphabetical, Equal Contribution)}
\contribution[\circ]{Intern}
\contribution[\dagger]{Project Lead}
\contribution[\S]{Equal Contribution}
\abstract{
We present SAM 3D, a generative model for visually grounded 3D object reconstruction, predicting geometry, texture, and layout from a single image. 
SAM 3D excels in natural images, where occlusion and scene clutter are common and visual recognition cues from context play a larger role.
We achieve this with a human- and model-in-the-loop pipeline for annotating object shape, texture, and pose, providing visually grounded 3D reconstruction data at unprecedented scale.  We learn from this data in a modern, multi-stage training framework that combines synthetic pretraining with real-world alignment, breaking the 3D ``data barrier''. 
We obtain significant gains over recent work, with at least a $5:1$ win rate in human preference tests on real-world objects and scenes.
We will release our code and model weights, an online demo, and a new challenging benchmark for in-the-wild 3D object reconstruction.
}

\metadata[Code]{\url{https://github.com/facebookresearch/sam-3d-objects}}
\metadata[Demo]{\url{https://www.aidemos.meta.com/segment-anything/editor/convert-image-to-3d}}
\metadata[Website]{\url{https://ai.meta.com/sam3d}}

\begin{document}
\maketitle

\begin{figure}[h!]
\centering
\vspace{5mm}
\includegraphics[width=\linewidth]{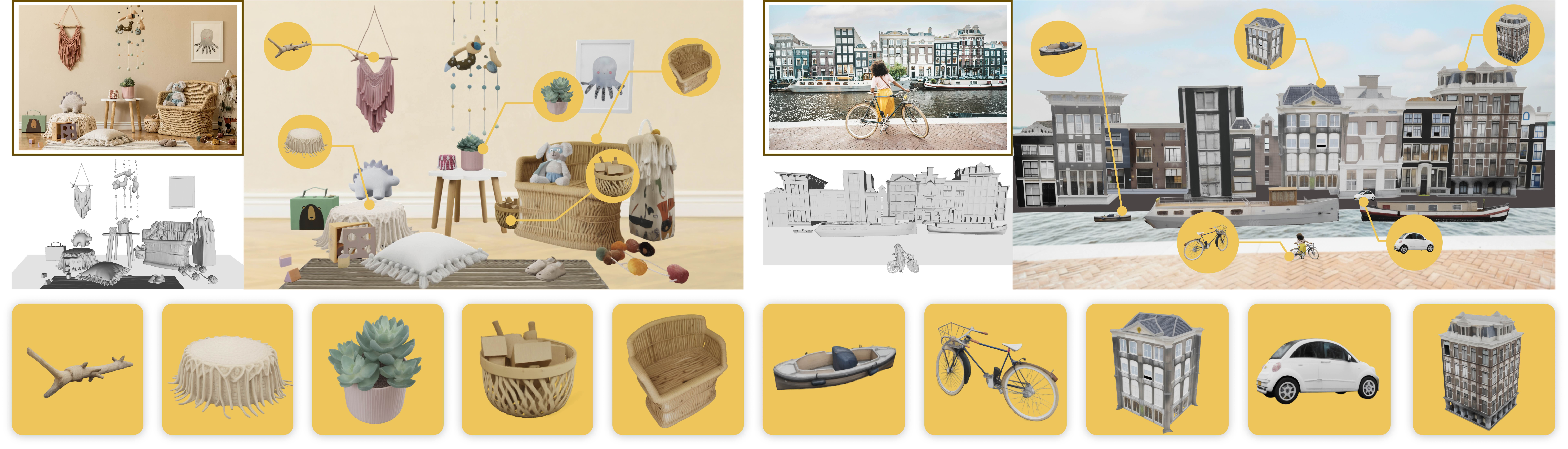}
\caption{
\textbf{\method converts a single image into a composable 3D scene made of individual objects.} Our method predicts per-object geometry, texture, and layout, enabling full scene reconstruction. Bottom: high-quality 3D assets recovered for each object.
}
\label{fig:teaser}
\end{figure}

\section{Introduction}
\label{section:intro}

In this paper (see \Cref{fig:teaser}) we present SAM 3D, a generative neural network for 3D  reconstruction from a single image. The model can reconstruct 3D shape and texture for any object, as well as its layout with respect to the camera, even in complex scenes with significant clutter and  occlusion. As the reconstruction is of full 3D shape, not just of the visible 2.5D surface, one can then re-render the object from any desired viewpoint.

Computer vision has traditionally focused on multi-view geometry as providing the primary signal for 3D shape. However psychologists (and artists before them) have long known that \emph{humans} can perceive depth and shape from a single image, \eg~\citet{koenderink1992surfacep} demonstrated this elegantly by showing that humans can estimate surface normals at probe points on an object's image, which can then be integrated to a full surface. In psychology textbooks these single image cues to 3D shape are called “pictorial cues”, and include information such as in shading and texture patterns, but also recognition - the  ``familiar object'' cue.  In computer vision, this line of research dates back to \citet{roberts1963machine}, who showed that once an image pattern was recognized as a known object, its 3D shape and pose could be recovered. The central insight is that  recognition enables 3D reconstruction, an idea that has since resurfaced in different technical instantiations~\citep{debevec2023modeling,cashman2012shape,kar2015category,gkioxari2019mesh,xiang2025structured,wu2024reconstructing}.  Note that this permits generalization to novel objects, because even if a specific object has not been seen before, it is made up of parts seen before. 

A fundamental challenge for learning such models is the lack of data: specifically, natural images paired with 3D ground truth are difficult to obtain at scale.
Recent work \citep{yang2024hunyuan3d, xiang2025structured} has shown strong reconstruction from single images. However, these models are trained on isolated objects and struggle with objects in natural scenes, where they may be distant or heavily occluded. To add such images to the training set, we need to find a way to associate specific objects in such images with 3D shape models, acknowledging that generalist human annotators find it hard to do so (unlike, say, attaching a label like ``cat'' or marking its boundary). Two insights made this possible:
\begin{itemize}
    \item We can create synthetic scenes where 3D object models are rendered and pasted into images (inspired by~\citet{dosovitskiy2015flownet}).
    \item While humans can't easily {\em generate} 3D shape models for objects, they can {\em select} the likely best 3D model from a set of proffered choices and align its pose to the image (or declare that none of the choices is good).
\end{itemize}

We design a training pipeline and data engine by adapting modern, multistage training recipes pioneered by LLMs \citep{minaee2025largelanguagemodelssurvey, mo2025midtraininglargelanguagemodels}.  As in recent works, we first train on a large collection of rendered synthetic objects. 
This is supervised pretraining: our model learns a rich vocabulary for object shape and texture, preparing it for real-world reconstruction. Next is mid-training with semi-synthetic data produced by pasting rendered models into natural images.  Finally, post-training adapts the model to real images, using both a novel model-in-the-loop (MITL) pipeline and human 3D artists, and aligns it to human preference.  We find that synthetic pretraining  generalizes, given adequate post-training on natural images.

Our post-training data, obtained from our MITL data pipeline, is key to obtaining good performance in natural images.  Generalist human annotators aren't capable of producing 3D shape ground truth; hence our annotators select and align 3D models to objects in images from the output of modules -- computational and retrieval-based -- that produce multiple initial 3D shape proposals. 
Human annotators select from these proposals, or route them to human artists for a subset of hard instances. The vetted annotations feed back into model training, and the improved model is reintegrated into the data engine to further boost annotation quality. This virtuous cycle steadily improves the quality of 3D annotations, labeling rates, and model performance.

Due to the lack of prior benchmarks for real-world 3D reconstruction of object shape and layout, we propose a new evaluation set of $1,000$ image and 3D pairs: SAM 3D Artist Objects (SA-3DAO). The objects in our benchmark range from churches, ski lifts, and large structures to animals, everyday household items, and rare objects, and are paired with the real-world images in which they naturally appear. Professional 3D artists create 3D shapes from the input image, representing an expert human upper bound for visually grounded 3D reconstruction. 
We hope that contributing such an evaluation benchmark helps accelerate subsequent research iteration of real-world 3D reconstruction models.

\noindent We summarize our contributions as follows:
\begin{itemize}
    \item We introduce \textbf{\method}, a new foundation model for 3D that predicts object shape, texture, and pose from a single image. 
    By releasing code, model weights, and a demo, we hope to stimulate further advancements in 3D reconstruction and downstream applications of 3D.
    \item We build a MITL pipeline for annotating shape, texture, and pose data, providing visually grounded 3D reconstruction data at unprecedented scale.
    \item We exploit this data via LLM-style pretraining and post-training in a novel framework for 3D reconstruction, combining synthetic pretraining with real-world alignment to overcome the orders of magnitude data gap between 3D and domains such as text, images, or video.
    \item We release a challenging benchmark for real-world 3D object reconstruction, SA-3DAO. Experiments show SAM 3D's significant gains via metrics and large-scale human preference.
\end{itemize}

\section{The \method Model}
\label{section:model}
\begin{figure*}[t]
    \centering 
    \includegraphics[width=0.9\linewidth]{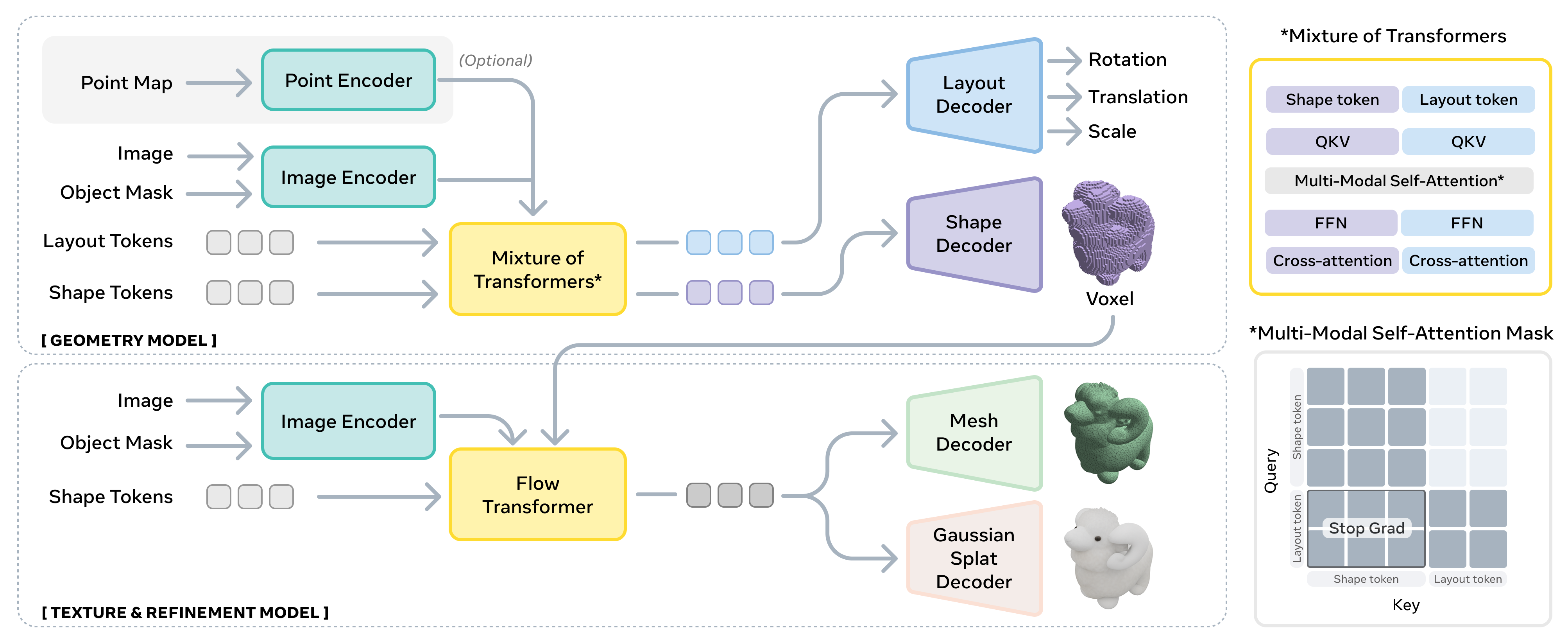}
    \caption{\textbf{\method architecture.} 
    (\textbf{top}) \method first predicts coarse shape and layout with the Geometry model; 
    (\textbf{right}) the mixture of transformers architecture apply a two-stream approach with information sharing in the multi-modal self-attention layer.
    (\textbf{bottom}) The voxels predicted by the Geometry model are passed to the Texture \& Refinement model, which adds higher resolution detail and textures.
    }
    \label{fig:method}
\end{figure*}

\subsection{Problem Formulation}
\label{sec:problem_formulation}
The act of taking a photograph maps a 3D object to a set of 2D pixels, specified by a mask $M$ in an image $I$.
We seek to invert this map.
Let the object have shape $S$, texture $T$, and rotation, translation and scale $(R, t, s)$ in camera coordinates.
 Since the 3D to 2D map is lossy, we  model the reconstruction problem as a conditional distribution $p(S, T, R, t, s | I, M)$.
Our goal is to train a generative model $q(S, T, R, t, s | I, M)$ that approximates $p$ as closely as possible.

\subsection{Architecture}
\label{sec:architecture}
We build upon recent SOTA two-stage latent flow matching architectures~\citep{xiang2025structured}.
\method first jointly predicts object pose and coarse shape, then refines the shapes by integrating pictorial cues (see~\Cref{fig:method}). 
Unlike~\citet{xiang2025structured} that reconstructs isolated objects, \method predicts object layout, creating coherent multi-object scenes.

\paragraph{Input encoding.}
We use  DINOv2~\citep{oquab2023dinov2} as an encoder to extract features from two pairs of images, resulting in 4 sets of conditioning tokens:
\begin{itemize}
    \item \textbf{Cropped object}: We encode the cropped image $I$ by mask $M$ and its corresponding \textit{cropped binary mask}, providing a focused, high-resolution view of the object.
    \item \textbf{Full image}: We encode the full image $I$ and its \textit{full image binary mask}, providing global scene context and recognition cues absent from the cropped view.
\end{itemize}

Optionally, the model supports conditioning on a coarse scene point map, $P$ obtained via hardware sensors (\eg, LiDAR on an iPhone), or monocular depth estimation~\citep{yang2024depth,wang2025moge}, enabling \method to integrate with other pipelines. 

\begin{figure*}
    \centering 
    \includegraphics[width=\linewidth]{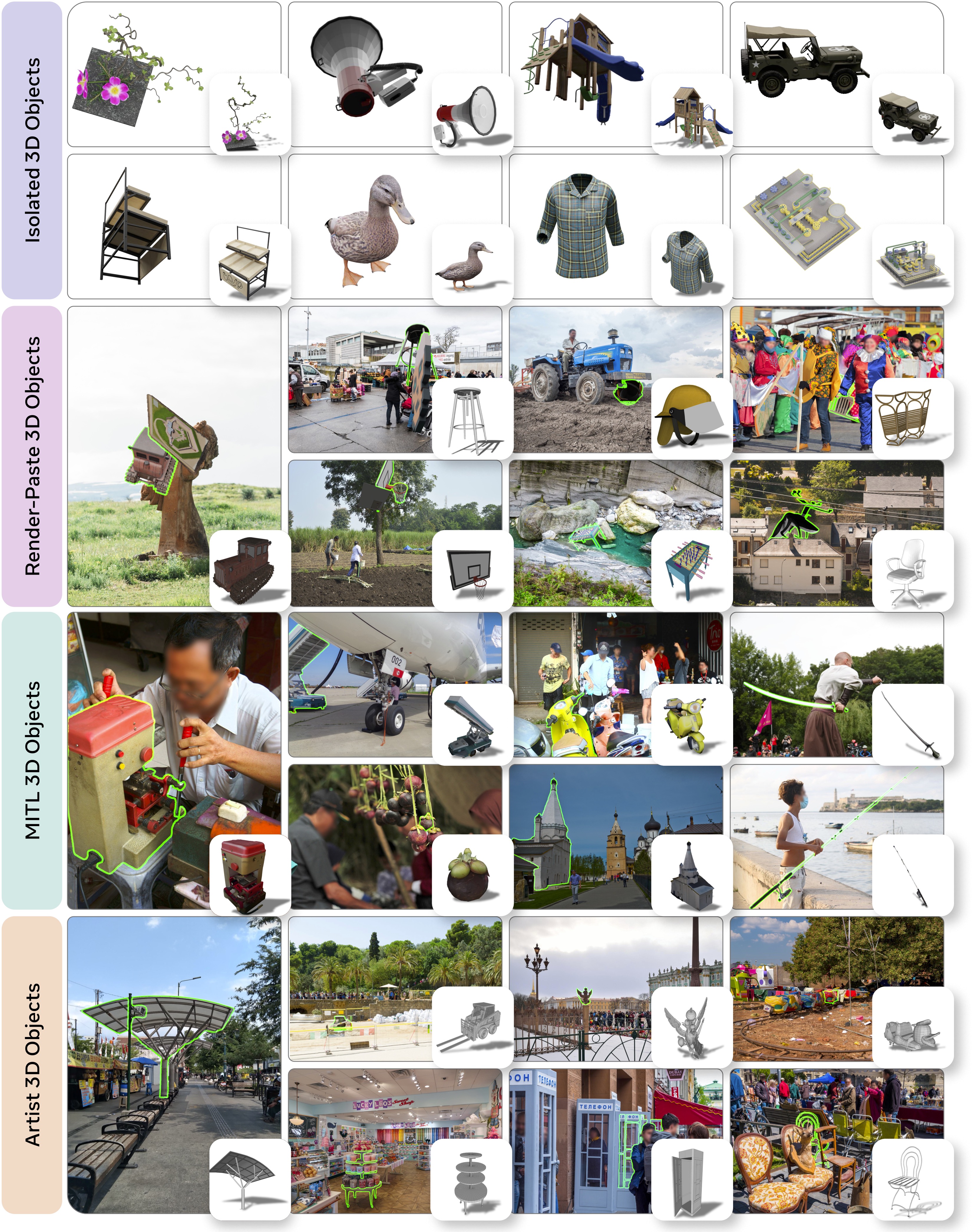}  
    \caption{
    \textbf{\method data,} with a green outline around the target object, and the ground truth mesh shown in the bottom right.
    Samples are divided into four rows, based on type.
    Art-3DO meshes are untextured, while the rest may be textured or not, depending on the underlying asset (Iso-3DO, RP-3DO) or if the mesh was annotated for texture (MITL-3DO).
    }
    \label{fig:training_data}
\end{figure*}

\paragraph{The Geometry Model} models the conditional distribution $p(O, R, t, s | I, M)$, where $O \in \mathbb{R}^{64^3}$ is coarse shape, $R \in \mathbb{R}^{6}$ the 6D rotation~\citep{zhou2019continuity}, $t \in \mathbb{R}^{3}$ the translation, and $s \in \mathbb{R}^{3}$ the scale.
Conditioned on the input image and mask encodings, we employ a 1.2B parameter flow transformer with the Mixture-of-Transformers (MoT) architecture~\citep{liang2025mixtureoftransformers,deng2025Bagel}, modeling geometry $O$ and layout $(R, t, s)$ using the attention mask in~\Cref{fig:method}. See~\Cref{sec:fm_details} for details.

\paragraph{The Texture \& Refinement Model} 
learns the conditional distribution $p(S, T | I, M, O)$. 
We first extract active voxels from the coarse shape $O$ predicted by Geometry model.
A 600M parameter sparse latent flow transformer~\citep{xiang2025structured,peebles2023scalable} refines geometric details and synthesizes object texture.

\paragraph{3D Decoders.} 
The latent representations from the Texture \& Refinement Model can be decoded to either mesh or 3D Gaussian splats via a pair of VAE decoders $\mathcal D_m$, $\mathcal D_g$.
These separately-trained decoders share the same VAE encoder and hence the same structured latent space~\citep{xiang2025structured}. We also detail several improvements in~\Cref{sec:vae_improvements}.

\section{Training \method}
\label{sec:training}
\method breaks the 3D data barrier using a recipe that progresses from synthetic pretraining to natural post-training, adapting the playbook from LLMs, robotics, and other large generative models. We build capabilities by stacking different training strategies in pre- and mid-training, and then align the model to real data and human-preferred behaviors through a post-training data flywheel.
\method uses the following approach:

\noindent\textbf{Step 1: Pretraining.} This phase builds foundational capabilities, such as shape generation, into a base model.

\noindent\textbf{Step 1.5: Mid-Training.} Sometimes called continued pretraining, mid-training imparts general skills such as occlusion robustness, mask-following, and using visual cues. 

\noindent\textbf{Step 2: Post-Training.} Post-training elicits target behavior, such as adapting the model from synthetic to real-world data or following human aesthetic preferences.
We collect training samples $(I,M)\rightarrow (S,T,R,t,s)$ and preference data from humans and use them in both supervised finetuning (SFT) and direct preference optimization (DPO)~\citep{rafailov2023direct}.

This alignment (step 2) can be repeated, first collecting data with the current model and then improving the model with the new data. This creates a virtuous cycle with humans providing the supervision. \Cref{fig:elo_data_engine} shows that as we run the data engine longer, model performance steadily improves; dataset generation emerges as a byproduct of this alignment.

The following sections detail the training objectives and data sources used in \method. We focus on the Geometry model;  Texture \& Refinement is trained similarly (details  in~\Cref{app:texture_training}). Training hyper-parameters are in~\Cref{sec:training_details}.

\begin{figure*}[t]
    \centering
    \includegraphics[width=\textwidth]{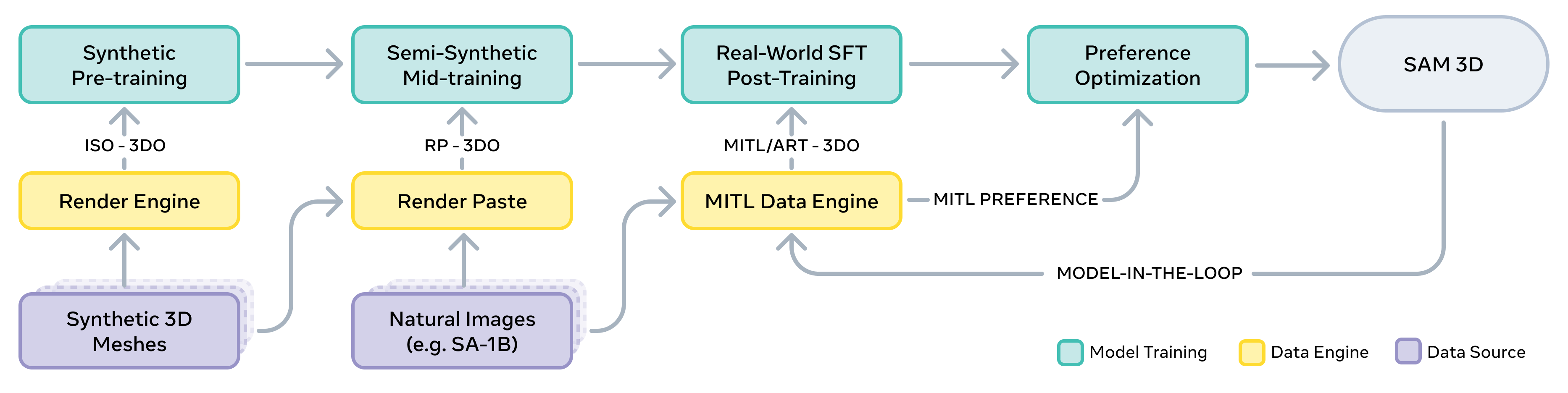}
    \vspace{-3mm}
    \caption{\textbf{\method training paradigm.}
    We employ a multi-stage pipeline incrementally exposing the model to increasingly complex data and modalities.}
    \label{fig:paradigm}
\end{figure*}

\begin{table*}[t]
    \centering
    \vspace{1mm}
    \begin{tabular*}{0.8\textwidth}{@{\extracolsep{\fill}}llll}
        \toprule
        \textbf{Training stage} & \textbf{Modalities} & \textbf{Datasets} & \textbf{Condition input} \\
        \midrule
        \multicolumn{4}{l}{\textbf{Stage 1 Geometry model}} \\
        \midrule
        Pre-training & $S, R$ & Iso-3DO & object-centric crop \\
        \multirow{2}{*}{Mid-training} 
            & $S, R$ & RP-3DO$^\dagger$ & full image \\
            & $S, R, t, s$ & ProcThor, RP-3DO$^\ddagger$ & full image, pointmap$^*$ \\
        SFT & $S, R, t, s$ & MITL, Art-3DO & full image, pointmap$^*$ \\
        Alignment & $S$ & MITL preference & full image, pointmap$^*$ \\
        \midrule
        \multicolumn{4}{l}{\textbf{Stage 2 Texture \& Refinement model}} \\
        \midrule
        Pre-training & $T$ & Iso-3DO-500K & object-centric crop \\
        Mid-training & $T$ & RP-3DO$^{\dagger,\S}$ & full image \\
        SFT & $T$ & MITL & full image \\
        Alignment & $T$ & MITL preference & full image \\
        \bottomrule
    \end{tabular*}
    \vspace{1mm}
    \caption{
        \textbf{\method training stages.}
        {\footnotesize
        $^\dagger$Flying Occlusion (FO) from RP-3DO. 
        $^\ddagger$Object Swap - Random (OS-R) from RP-3DO. 
        $^\S$Object Swap - Annotated (OS-A) from RP-3DO. 
        $^*$optional.
        See~\Cref{app:mid_training_data} for details.
        }
    }
    \label{table:training_diagram}
\end{table*}

\subsection{Pre \& Mid-Training: Building a Base Model}
\label{section:pretraining_data}

Training begins with synthetic pretraining and mid-training, leveraging available large-scale datasets to learn strong priors for shape and texture, and skills such as mask-following, occlusion handling, and pose estimation. The rich features learned here drastically reduce the number of labeled real-world samples required in post-training~\citep{hernandez2021scaling}, which generally incur acquisition costs.
In pre- and mid-training, models are trained with rectified conditional flow matching~\citep{liu2022flow} to generate multiple 3D modalities (see \Cref{sec:flow_matching_training_objective}).

\subsubsection{Pretraining: Single Isolated 3D Assets}
Pretraining trains the model to reconstruct accurate 3D shapes and textures from renders of isolated synthetic objects, following the successful recipes from~\citep{xiang2025structured,yang2024hunyuan3d,wu2024direct3d}. 
Specifically, we gather a set of image $I$, shape $S$, and texture $T$ triplets, using $2.7$ million object meshes from Objaverse-XL~\citep{deitke2023objaverse} and licensed datasets, and render them from $24$ viewpoints, each producing a high-resolution image of a single centered object; more detail in~\Cref{sec:iso_preprocessing}.
We call this dataset \emph{Iso-3DO} and train for 2.5 trillion training tokens.

\subsubsection{Mid-Training: Semi-Synthetic Capabilities}

Next, mid-training builds up foundational skills that will enable the model to handle objects in real-world images:
\begin{itemize}
    \item \textit{Mask-following}: 
     We train the model to reconstruct a target object, defined by a binary mask on the input image. 
    \item \textit{Occlusion robustness}: 
    The artificial occluders in our dataset incentivize learning shape completion.
    \item \textit{Layout estimation}: We train the model to produce translation and scale in normalized camera coordinates.
\end{itemize}

We construct our data by rendering textured meshes into natural images using alpha compositing.
This ``render-paste'' dataset contains one subset of occluder-occludee pairs, and another subset with real objects replaced by synthetic objects at similar location and scale, creating physically-plausible data with accurate 3D ground truth.  We call these $61$ million samples with $2.8$ million unique meshes \emph{RP-3DO}; \cref{fig:training_data} shows examples.
See~\Cref{app:mid_training_data} for more details.

After mid-training (2.7 trillion training tokens), the model has now been trained with all input and output modalities for visually grounded 3D reconstruction.
However, all data used has been (semi-)synthetic; to both close the domain gap and fully leverage real-world cues, we need real images.

\begin{figure*}[t]
    \centering \includegraphics[width=1.0\linewidth]{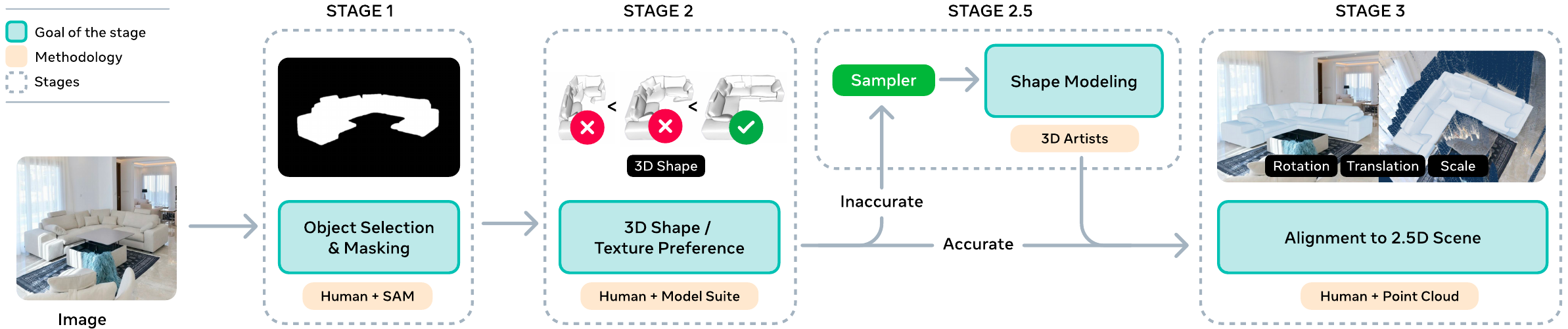}
    \caption{\textbf{Life of an example going through the data collection pipeline.} 
    We streamline annotation by breaking it into subtasks: annotators first choose target objects (Stage 1); rank and select 3D model candidates (Stage 2); then pose these models within a 2.5D scene (Stage 3). Stages 2 and 3 use model-in-the-loop.
    }
    \label{fig:data_engine}
\end{figure*}

\subsection{Post-Training: Real-World Alignment}
\label{sec:data_engine_alg}
In post-training, we have two goals.  The first is to close the domain gap between (semi-)synthetic data and natural images.  The second is to align with human preference for shape quality.
We adapt the model by using our data engine iteratively; we first \textbf{(i) collect training data} with the current model, and then \textbf{(ii) update our model} using multi-stage post-training on this collected data.  We then repeat.

\subsubsection{Post-Training: Collection Step}
\label{section:data}

The core challenge with collecting data for 3D visual grounding is that most people cannot create meshes directly; this requires skilled 3D artists, who even then can take multiple hours. This is different from the segmentation masks collected in SAM~\citep{kirillov2023segment}.
However, given options, most people \emph{can} choose which mesh resembles an object in the image most accurately. This fact forms the foundation of our data collection for \method.
We convert preferences into training data as follows: sample from our post-trained model, ask annotators to choose the best candidate and then grade its overall quality according to a rubric which we define and update. If the quality meets the (evolving) bar, the candidate becomes a training sample.

Unfortunately at the first iteration, our initial model yields few high-quality candidates. This is because before the first collection step, very little real-world data for 3D visual grounding exists. We deal with this cold start problem by leveraging a suite of existing learned and retrieval-based models to produce candidates.  In early stages, we draw mostly from the ensemble, but as training progresses our best model dominates, eventually producing about $80\%$ of the annotated data seen by \method.

Our annotation pipeline collects 3D object shape $S$, texture $T$, orientation $R$, 3D location $t$, and scale $s$ from real-world images.
We streamline the process by dividing into subtasks and leveraging existing appropriate models and human annotators within each (see \Cref{fig:data_engine}): identifying target objects, 3D model ranking and selection, and posing these within a 3D scene (relative to a point map).
We outline each stage of the data engine below and present details in~\Cref{sec:data_eng_details}. 
In total, we annotate almost $1$ million images with $\sim3.14$ million untextured meshes and $\sim100$K textured meshes--unprecedented scale for 3D data paired with natural images.

\paragraph{Stage 1: Choosing target objects $(I,M)$.}
The goal of this stage is to identify a large, diverse set of images $I$ and object masks $M$ to lift to 3D. 
To ensure generalization across objects and scenes, we sample images from several diverse real-world datasets, and utilize a 3D-oriented taxonomy to balance the object distribution.
To obtain object segmentation masks, we use a combination of pre-existing annotations~\citep{kirillov2023segment} and human labelers selecting objects of interest.

\begin{figure*}[t!!]
    \centering 
    \includegraphics[width=0.95\linewidth]{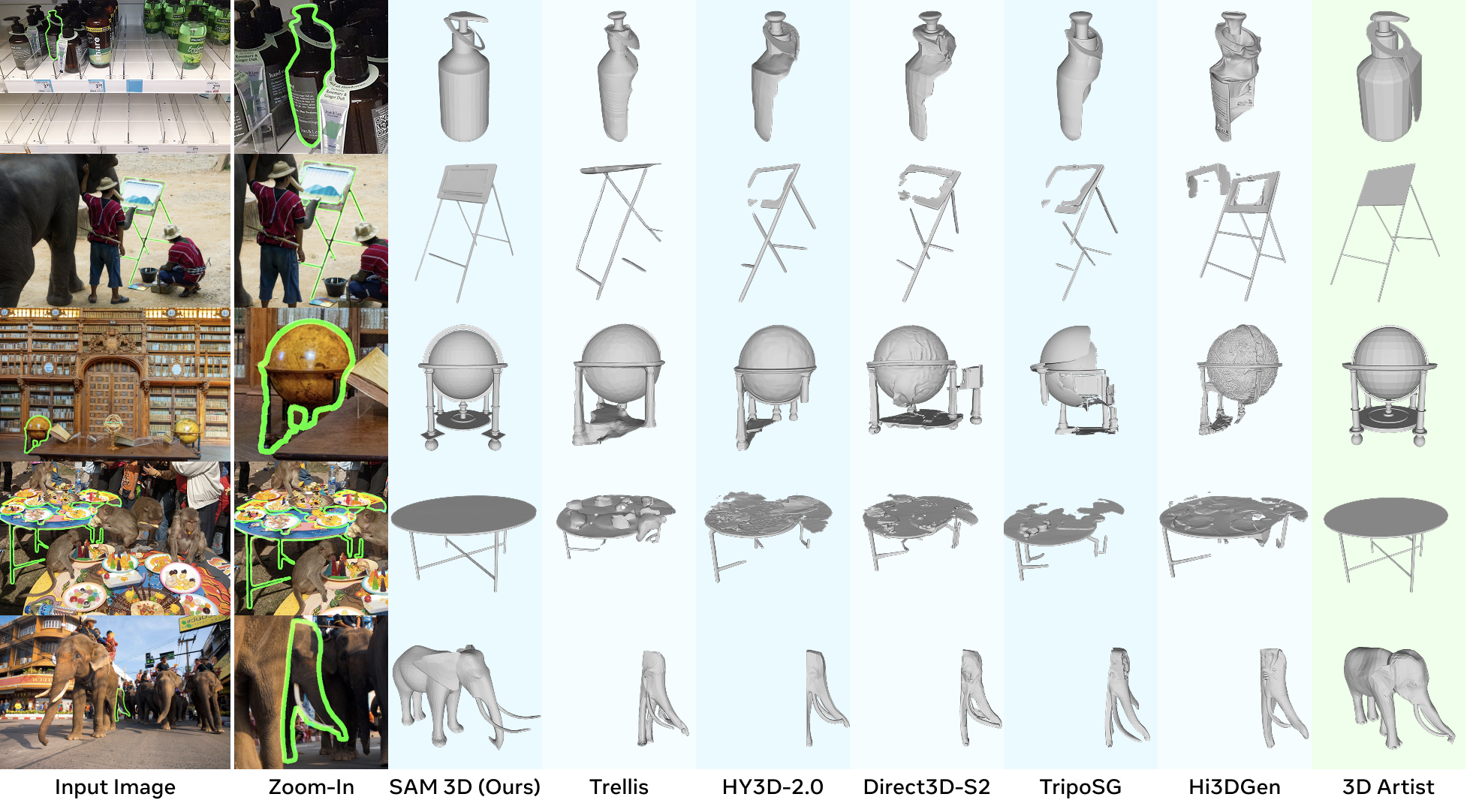}
    \caption{\textbf{Qualitative comparison to competing image-to-3D asset methods.} We compare to the recent Trellis~\citep{xiang2025structured}, Hunyuan3D-2.1~\citep{hunyuan3d2025hunyuan3d}, Direct3D-S2~\citep{wu2025direct3ds2gigascale3dgeneration} and Hi3DGen~\citep{ye2025hi3dgen} on the artist-generated SA-3DAO for single shape reconstruction; we provide the 3D artist-created ground truth mesh as reference.
    }
    \label{fig:comparison_shape}
    \vspace{-2mm}
\end{figure*}

\paragraph{Stage 2: Object model ranking and selection $(S,T)$.}
The goal of this stage is to collect image-grounded 3D shape $S$ and texture $T$.  As described above, human annotators choose shape and texture candidates which best match the input image and mask. Annotators rate the example $r$ and reject chosen examples that do not meet a predefined quality threshold, \ie $r < \alpha$. Bad candidates also become negative examples for preference alignment. 

Our data engine maximizes the likelihood of a successful annotation, $r > \alpha$, by asking annotators to choose between $N=8$ candidates from the ensemble; a form of best-of-$N$ search~\citep{ouyang2022training} using humans. The expected quality of this best candidate improves with $N$, and we further increase $N$ by first filtering using a model, and then filtering using the human~\citep{thomas2017exit}; we show results in~\Cref{sec:test_time_search}. 

\paragraph{Stage 2.5: Hard example triage (Artists).}
When no model produces a reasonable object shape, our non-specialist annotators cannot correct the meshes, resulting in a lack of data precisely where the model needs it most.  We route a small percentage of these hardest cases to professional 3D artists for direct annotation, and we denote this set \emph{Art-3DO}. 

\begin{figure*}[t!!]
    \centering 
    \includegraphics[width=0.95\linewidth]{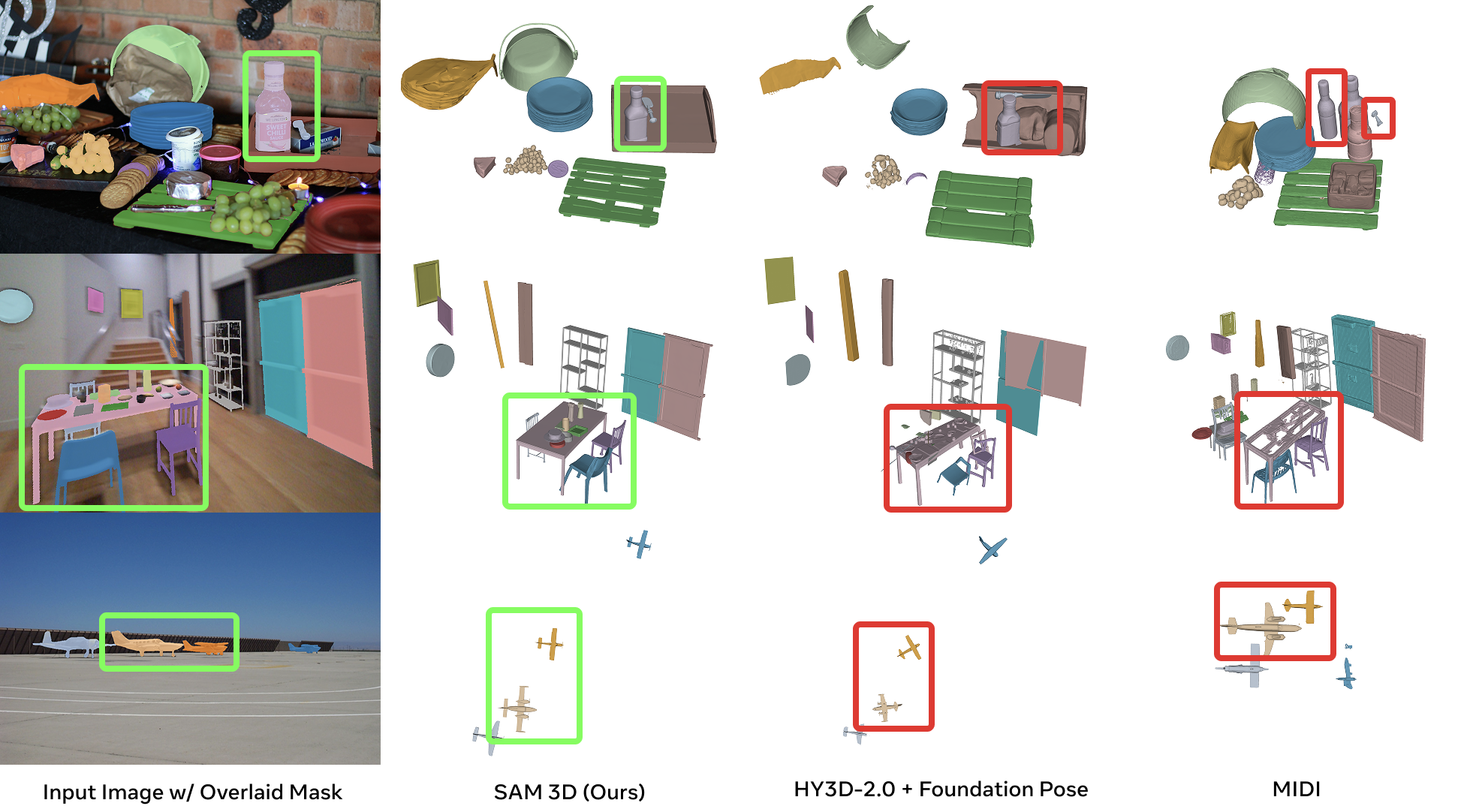}
    \caption{\textbf{Qualitative comparison to competing scene reconstruction methods.} We show SAM 3D's full 3D scene reconstructions versus alternatives~\citep{wen2024foundationpose,huang2025midi}.}
    \label{fig:comparison_scene}
    \vspace{-2mm}
\end{figure*}

\paragraph{Stage 3: Aligning objects to 2.5D scene $(R,t,s)$.}
The previous stages produce a 3D shape for the object, but not its layout in the scene.
For each stage 2 shape, annotators label the object pose by manipulating the 3D object's translation, rotation, and scale relative to a point cloud.
We find that point clouds provide enough structure to enable consistent shape placement and orientation. 

In general, we can think of the data collection as an API that takes a current best model, $q(S,T,R,t,s~|~I,M)$, and returns (i) training samples $D^{+} = (I,M,S,T,R,t,s)$, (ii) a quality rating $r\in[0,1]$, and (iii) a set of less preferred candidates ($D^{-} = (I,M,S',T',R',t',s')$) that are all worse than the training sample.

\subsubsection{Post-Training: Model Improvement Step}
The model improvement step in \method uses these training samples and preference results to update the base model through multiple stages of finetuning and preference alignment. 
Within each post-training iteration we aggregate data from all previous collection steps; keeping only samples where $D^{+}$ is above some quality threshold $\alpha$.  As training progresses, $\alpha$ can increase over time, similar to the cross-entropy method for optimization~\citep{2005crossentropymethod}. Our final post-training iteration uses 0.5 trillion training tokens.

\paragraph{Supervised Fine-Tuning (SFT).}
When post-training begins, the base model has only seen synthetic data.  Due to the large domain gap between synthetic and real-world data, we begin by finetuning on our aligned meshes from Stage 3.

We begin SFT with the noisier non-expert labels (MITL-3DO), followed by the smaller, high-quality set from 3D artists (Art-3DO).
The high quality Art-3DO data enhances model quality by aligning outputs with artists' aesthetic preferences. We find this helps suppress common failures, \eg floaters, bottomless meshes, and missing symmetry.

\paragraph{Preference optimization (alignment).}
\label{sec:dpo}
After fine-tuning, the model can robustly generate shape and layout for diverse objects and real-world images.
However, humans are sensitive to properties like symmetry, closure, etc. which are difficult to capture with generic objectives like flow matching.
Thus, we follow SFT with a stage of direct preference optimization (DPO)~\citep{rafailov2023direct}, using $D^{+}$/$D^{-}$ pairs from Stage 2 of our data engine.
We found this off-policy data was effective at eliminating undesirable model outputs, even after SFT on Art-3DO. DPO training details are in \Cref{sec:dpo_training_objective}.

\paragraph{Distillation.}
Finally, to enable sub-second shape and layout from the Geometry model, we finish a short distillation stage to reduce the number of function evaluations (NFE) required during inference from $25\rightarrow4$. We adapt~\citet{frans2024one} to our setting, and describe the details in~\Cref{sec:inference}.

\section{Experiments}
\label{section:results}

\begin{table}[t]
    \centering
        \begin{NiceTabular}{lcccccc}
        \CodeBefore
        \Body
        \toprule
            & \multicolumn{4}{c}{SA-3DAO} & \multicolumn{2}{c}{ISO3D Eval Set} \\
        \cmidrule(lr){2-5} \cmidrule(lr){6-7}
        Model 
            & F1@0.01 ($\uparrow$) 
            & vIoU ($\uparrow$) 
            & Chamfer ($\downarrow$)  
            & EMD ($\downarrow$)
            & ULIP ($\uparrow$) 
            & Uni3D ($\uparrow$)  \\
            
        \midrule
         Trellis
            & 0.1475 & 0.1392 & 0.0902 & 0.2131
            & 0.1473 & 0.3698\\
         HY3D-2.1
            & 0.1399 & 0.1266 & 0.1126 & 0.2432
            & 0.1293 & 0.3546 \\
         HY3D-2.0
            & 0.1574 & 0.1504 & 0.0866 & 0.2049
            & 0.1484 & 0.3662 \\
         Direct3D-S2 
            & 0.1513 & 0.1465 & 0.0962 & 0.2160
            & 0.1405 & 0.3653 \\
         TripoSG
            & 0.1533 & 0.1445 & 0.0844 & 0.2057 
            & \textbf{0.1529} & 0.3687 \\
         Hi3DGen
            & 0.1629 & 0.1531 & 0.0937 & 0.2134 
            & 0.1419 & 0.3594 \\
        \midrule 
        \textbf{\method} 
            & \textbf{0.2344} & \textbf{0.2311} & \textbf{0.0400} & \textbf{0.1211}
            & 0.1488 & \textbf{0.3707}\\
        \bottomrule
        \end{NiceTabular}
    \caption{\textbf{3D shape quantitative comparison} to competing image-to-3D methods, including Trellis~\citep{xiang2025structured}, HY3D-2.1~\citep{hunyuan3d2025hunyuan3d}, HY3D-2.0~\citep{hunyuan3d22025tencent}, Direct3D-S2~\citep{wu2025direct3ds2gigascale3dgeneration}, TripoSG~\citep{li2025triposg}, Hi3DGen~\citep{ye2025hi3dgen}.
    SA-3DAO shows metrics that measure accuracy against GT geometry; ISO3D~\citep{ebert20253d} has no geometric GT and so we show perceptual similarities between 3D and input images (ULIP~\citep{xue2023ulip} and Uni3D~\citep{zhou2023uni3d}). TripoSG uses a significantly higher mesh resolution, which is rewarded in perceptual metrics.
    }
    \label{table:comparison_shape}
\end{table}

\begin{figure}[t!]
    \centering
    \includegraphics[width=0.8\linewidth]{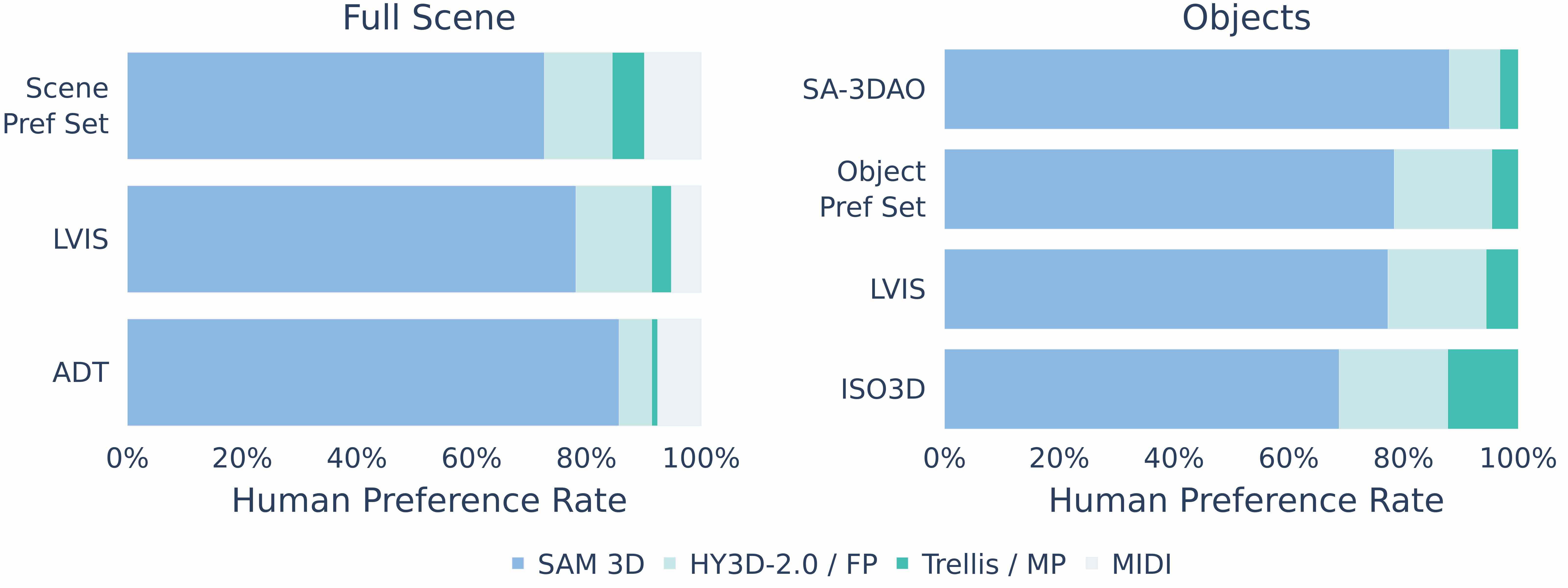}
    \caption{\textbf{Preference comparison on scene-level and object-level reconstruction.} Numbers indicate human preference rates. Objects comparisons are done on textured meshes. \method is significantly preferred over others on all fronts.}
    \label{fig:preference}
\end{figure}

\paragraph{Dataset.} To comprehensively evaluate the model capability under real-world scenarios, we carefully build a new benchmark \textbf{SA-3DAO}\footnote{\url{https://ai.meta.com/datasets/sa-3dao-sam-3d-artist-objects}}, consisting of 1K 3D artist-created meshes created from natural images. We also include \textbf{ISO3D} from 3D Arena~\citep{ebert20253d} for quantitatively evaluating shape and texture, and Aria Digital Twin (\textbf{ADT})~\citep{pan2023aria} for layout. We further conduct human preference evaluation on two curated sets for both scene-level and object-level reconstruction. The \textbf{Pref Set} uses real-world images from MetaCLIP~\citep{xu2024metaclip} and SA-1B~\citep{kirillov2023segment}, as well as a set based on LVIS~\citep{gupta2019lvis}. Refer to~\Cref{section:eval} for details on evaluation sets.

\paragraph{Settings.} 
We conduct experiments with a Geometry model trained to condition on pointmaps. For datasets where pointmaps are unavailable, we estimate them with~\citet{wang2025moge}. We found that shape and texture quality do not depend on whether the model is trained with pointmap conditioning (see \Cref{sec:rgb_only_model_ablation}), but layout (translation/scale) evaluation in~\Cref{table:comparison_layout} requires ground-truth depth/pointmap as reference.

\subsection{Comparison with SOTA}
\begin{table*}[t]
    \centering
    \resizebox{1.0\textwidth}{!}{
        \begin{NiceTabular}{clcccccccc}
        \CodeBefore
        \Body
        \toprule
            & &  \multicolumn{4}{c}{SA-3DAO} & \multicolumn{4}{c}{Aria Digital Twin} \\
        \cmidrule(lr){3-6} \cmidrule(lr){7-10}
        Generation & Model   & 3D IoU ($\uparrow$) & ICP-Rot. ($\downarrow$) & ADD-S ($\downarrow$) & ADD-S @ 0.1 ($\uparrow$) & 3D IoU ($\uparrow$) & ICP-Rot. ($\downarrow$) & ADD-S ($\downarrow$) & ADD-S @ 0.1 ($\uparrow$)   \\
        \midrule
         Pipeline & Trellis + Megapose & 0.2449 & 39.3866 & 0.5391 & 0.2831 & 0.2531 & 33.6114 & 0.4358 & 0.1971  \\         
         Pipeline & HY3D-2.0 + Megapose & 0.2518 & 33.8307 & 0.7146 & 0.3647 & 0.3794 & 29.0066 & 0.1457 & 0.4211  \\         
         Pipeline & HY3D-2.0 + FoundationPose & 0.2937 & 32.9444 & 0.3705 & 0.5396 & 0.3864 & 25.1435 & 0.1026 & 0.5992  \\
         Pipeline & HY3D-2.1 + FoundationPose & 0.2395 & 39.8357 & 0.4186 & 0.4177 & 0.2795 & 33.1197 & 0.2135 & 0.4129  \\
         Pipeline & \method + FoundationPose  & 0.2837 & 32.9168 & 0.3848 & 0.5079 & 0.3661 & 18.9102 & 0.0930 & 0.6495  \\
         \addlinespace[0.5em]
         Joint & MIDI & - & - & - & - & 0.0336 & 44.2353 & 2.5278 & 0.0175  \\         
         \midrule
         Joint & \textbf{\method} & \textbf{0.4254} & \textbf{20.7667} & \textbf{0.2661} & \textbf{0.7232} & \textbf{0.4970} & \textbf{15.2515} & \textbf{0.0765} & \textbf{0.7673} \\
        \bottomrule
        \end{NiceTabular}
    } 
    \caption{\textbf{3D layout quantitative comparison} to competing layout prediction methods on SA-3DAO and Aria Digital Twin~\citep{pan2023aria}. \method significantly outperforms both \emph{pipeline} approaches used in robotics~\citep{labbe2022megapose,wen2024foundationpose} and \emph{joint} generative models (MIDI~\citep{huang2025midi}). Most SA-3DAO scenes only contain one object so we do not show MIDI results that require multi-object alignment. The metrics measure bounding box overlap, rotation error, and chamfer-like distances normalized by object diameter.
    }
    \label{table:comparison_layout}
\end{table*}

\begin{figure}[t!]
    \centering
    \includegraphics[width=0.7\linewidth]{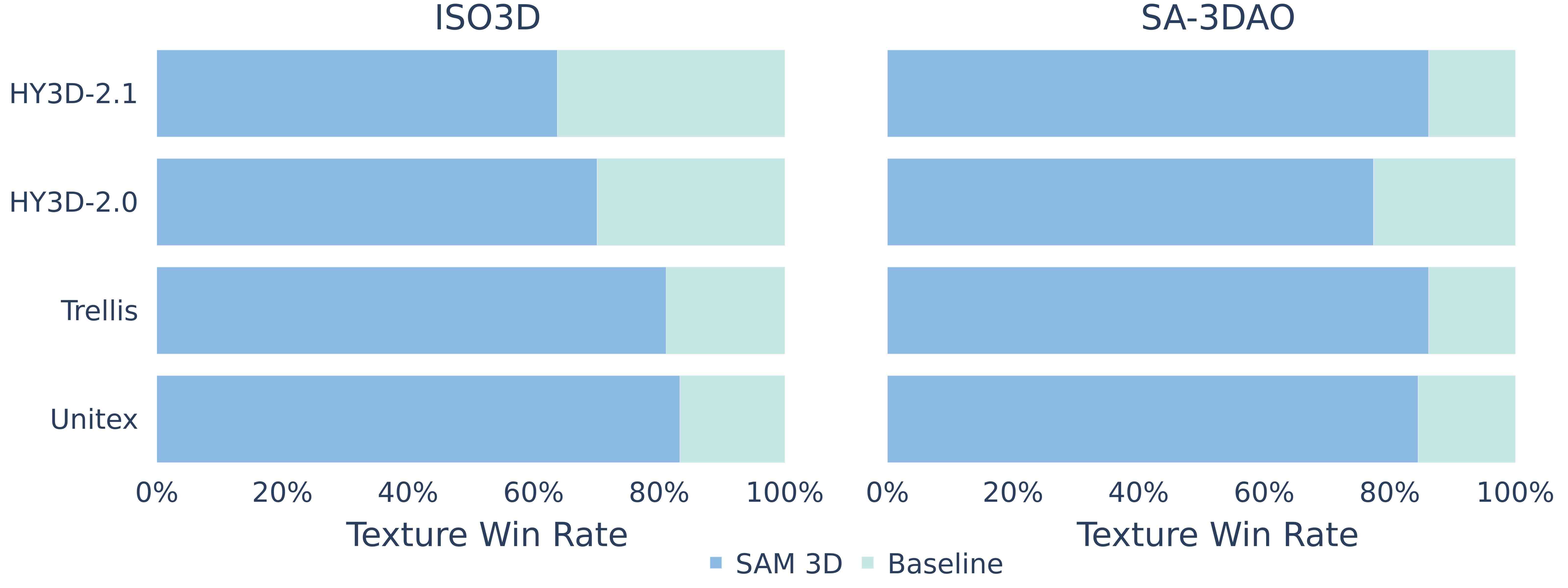}
    \caption{\textbf{Preference comparison on texture}. Since \method provides higher quality shape, we use the geometry results from \method and only perform texture generations for all methods. \method significantly outperforms others.}
    \label{fig:texture_preference}
\end{figure}

\paragraph{3D shape and texture.}
We evaluate single-object generation by comparing \method with prior state-of-the-art (SOTA) methods. In human preference studies, \method achieves a $5:1$ head-to-head win rate on real images (see~\Cref{fig:preference}). \Cref{table:comparison_shape} presents quantitative evaluation on shape quality, where \method matches or exceeds previous SOTA performance on isolated object images (\textbf{ISO3D}), and significantly outperforms all baselines on challenging real-world inputs (\textbf{SA-3DAO}). Qualitative examples in~\Cref{fig:comparison_shape} further illustrate the model’s strong generalization under heavy occlusion. In~\Cref{fig:texture_preference}, we compare \method texture vs. other texture models, given \method shapes (\method's improved shape actually benefits other methods in this eval).  Annotators significantly prefer \method texture (details in~\Cref{sec:texture_eval}).

\paragraph{3D scene reconstruction.}
In preference tests on three evaluation sets, users prefer scene reconstructions from \method by $6:1$ over prior SOTA (\Cref{fig:preference}). 
\Cref{fig:comparison_scene} and \Cref{fig:layout_visual} in the appendix show qualitative comparisons, while \Cref{table:comparison_layout} shows quantitative metrics for object layout. 
On real-world data like \textbf{SA-3DAO} and \textbf{ADT}, the improvement is fairly stark and persists even when \emph{pipeline} approaches use \method meshes.
\method introduces a new real-world capability to generate shape and layout \emph{jointly} (ADD-S @ 0.1 $2\% \rightarrow 77\%$), and a sample-then-optimize approach, as in the render-and-compare approaches~\citep{labbe2022megapose,wen2024foundationpose} can further improve performance (\Cref{sec:post-optim}).
The strong results for layout and scene reconstruction demonstrate that \method can robustly handle both RGB-only inputs (\eg, \textbf{SA-3DAO}, \textbf{LVIS}, \textbf{Pref Set}) as well as provided pointmaps (\eg, \textbf{ADT}).

\subsection{Analysis Studies}
\begin{figure}[t]
  \centering
  \begin{subfigure}[]{0.4\columnwidth}
    \centering
    \includegraphics[width=\textwidth]{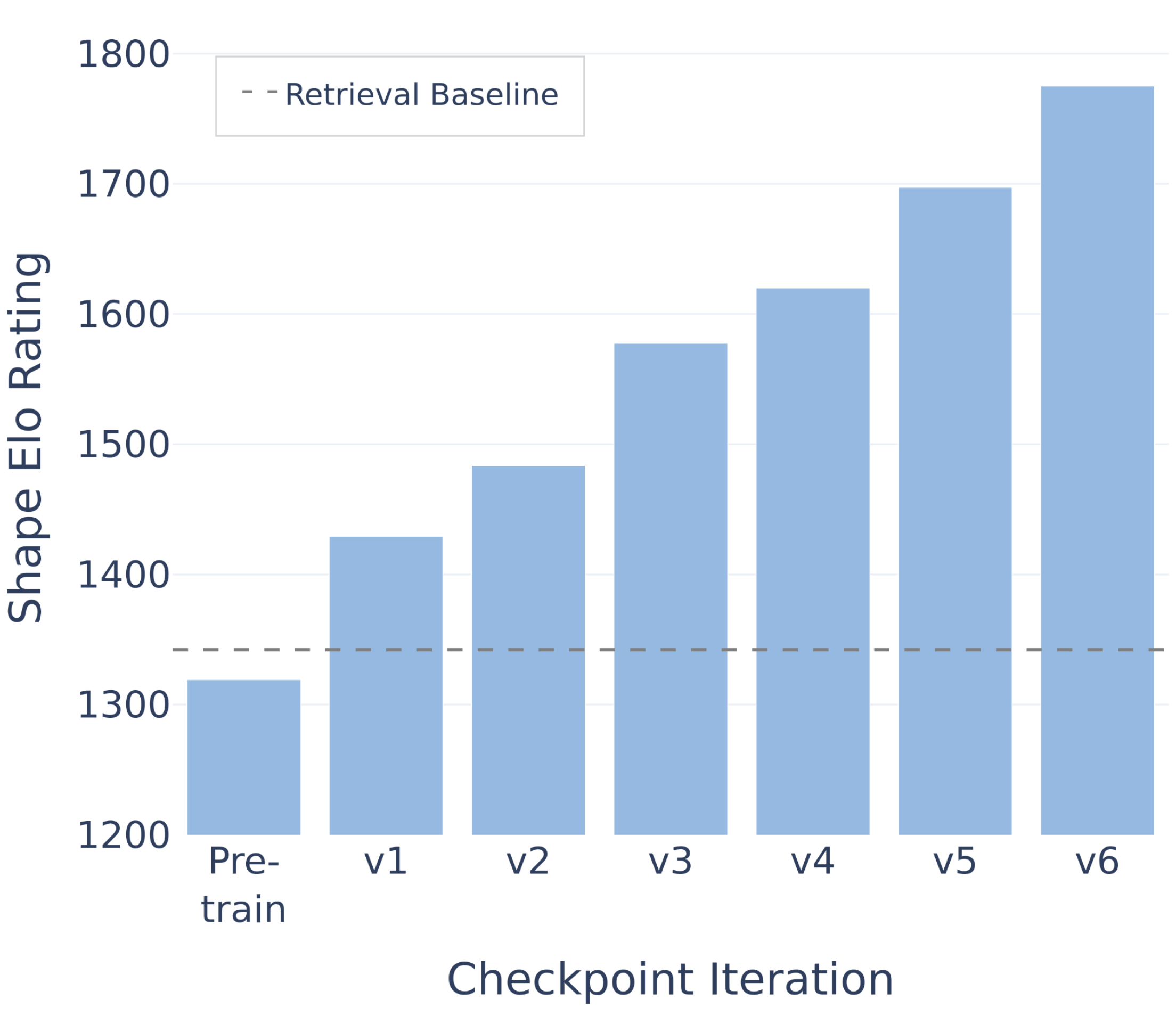}
    \caption{Historical Elo from data engine}
    \label{fig:elo_data_engine_historic}
  \end{subfigure}
  \hspace{10mm}
  \begin{subfigure}[]{0.4\columnwidth}
    \centering
    \includegraphics[width=\textwidth]{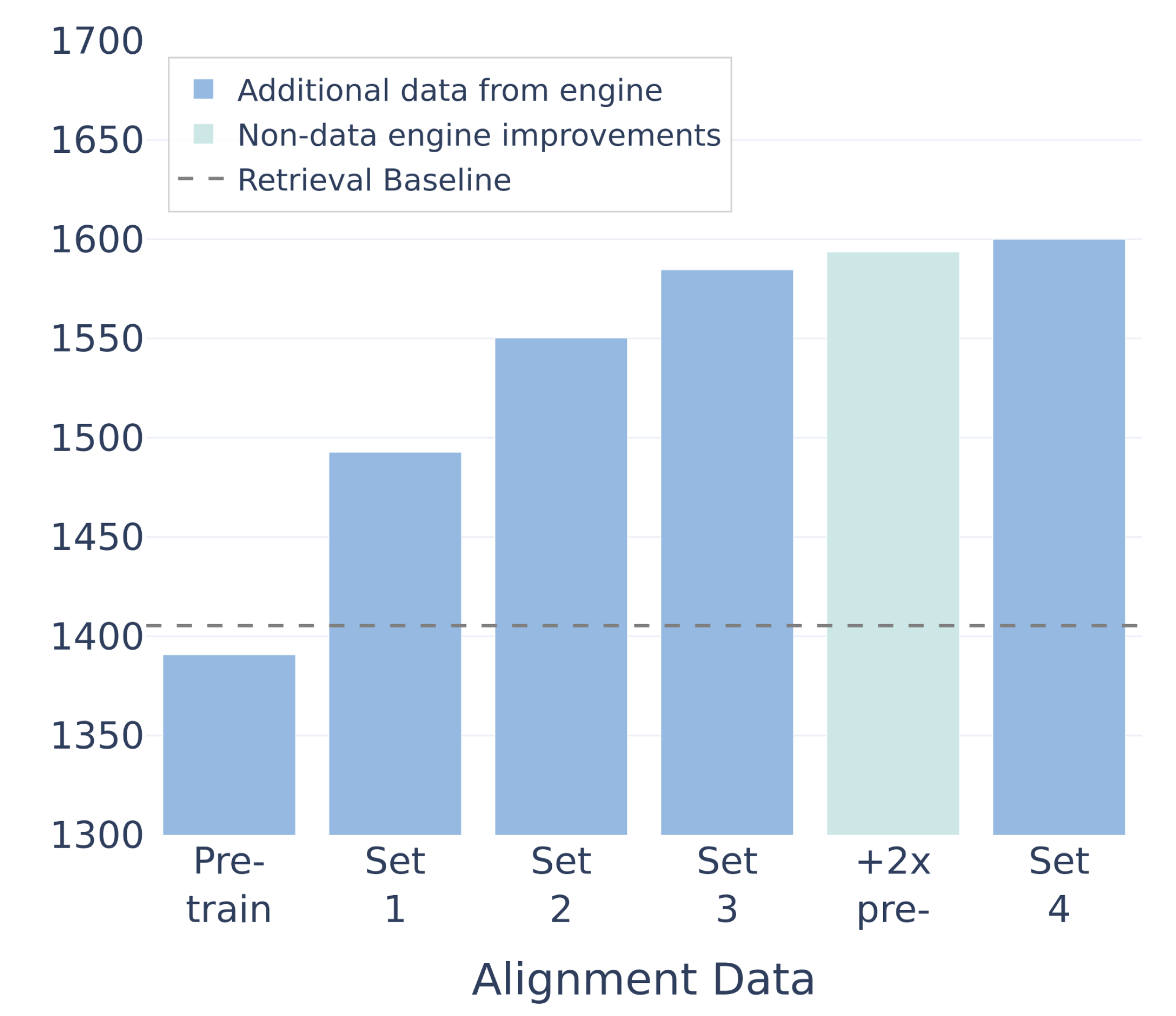}
    \caption{Impact of expanding training data}
    \label{fig:elo_data_engine}
  \end{subfigure}
  \hfill
  
  \caption{\textbf{Data engine with additional iterations.} The plots show Elo scores of different models; a 400 point Elo difference corresponds to 10:1 odds in a preference test. 
  Models were \textbf{(a)} checkpoints around 3 weeks apart, indicating cumulative improvements as we scale and add different stages and \textbf{(b)} post-trained (SFT) using expanded training data.}
  \label{fig:elo_combined}
\end{figure}

\paragraph{Post-training iterations steadily improve performance.}
We observed steady improvements as we ran the data engine for longer, 
with near-linear Elo scaling shown in the historical comparisons from Stage 2 of our data engine (\Cref{fig:elo_data_engine_historic}). We found it important to scale all stages simultaneously. The cumulatively linear effect results from more data engine iterations, along with scaling up pretraining, mid-training, and adding new post-training stages.
\Cref{fig:elo_data_engine} shows that iterating MITL-3DO data alone yields consistent improvements but with decreasing marginal impact. 

\begin{table}[t]
    \centering
        \begin{NiceTabular}{lccccc}
        \CodeBefore
        \Body
        \toprule
        & \multicolumn{4}{c}{SA-3DAO} & Preference set \\
        \cmidrule(lr){2-5} \cmidrule(lr){6-6}
        Training Stage   & F1 @ 0.01 ($\uparrow$) & vIoU ($\uparrow$) & Chamfer ($\downarrow$) & EMD ($\downarrow$)  & Texture WR ($\uparrow$) \\
        \midrule
        Pre-training (Iso-3DO) & 0.1349 & 0.1202 & 0.1036 & 0.2396 & - \\
        + Mid-training (RP-3DO) &  0.1705 & 0.1683 & 0.0760 & 0.1821 & 60.7 \\
        + SFT (MITL-3DO) & 0.2027 & 0.2025 & 0.0578 & 0.1510 & 66.9 \\
        + DPO (MITL-3DO) & 0.2156 & 0.2156 & 0.0498 & 0.1367 & 66.4 \\
        + SFT (Art-3DO)  & 0.2331 & \textbf{0.2337} & 0.0445 & 0.1257 & - \\
        + DPO (Art-3DO) & \textbf{0.2344} & 0.2311 & \textbf{0.0400} & \textbf{0.1211} & - \\
        \bottomrule
        \end{NiceTabular}
    \caption{\textbf{Cascading improvements from multi-stage training on 3D shape and texture}. 
    For texture, we report win rates (WR) between each row and the row \textit{above} it.}
    \label{table:stage}
\end{table}

\paragraph{Multi-stage training improves performance.}
\method's real-world performance emerges through multi-stage training.
\Cref{table:stage} reveals near-monotonic 3D shape improvements as each training stage is added, validating the approach that leads to the final model (last row). In the appendix, \Cref{fig:ablations_texture} shows similar results for texture and 
\Cref{table:ablate_stage} shows the contribution of each individual real-world data stage, by knocking out the MITL-3DO, Art-3DO data, or DPO stages.

\paragraph{Other ablations.}
Please see the appendix for additional ablations on rotation representation (\Cref{sec:rot_representation_ablation}), DPO (\Cref{sec:dpo_training_objective}), distillation (\Cref{sec:inference}), pointmaps (\Cref{sec:rgb_only_model_ablation}), and scaling best-of-$N$ in the data engine (\Cref{sec:test_time_search}).

\section{Related Work}
\label{section:related_work}

\paragraph{3D reconstruction} has been a longstanding challenge in computer vision.
Classical methods leverage multiple views and include binocular stereopsis~\citep{wheatstone1838xviii}, structure-from-motion~\citep{hartley2003multiple,szeliski2022computer,scharstein2002taxonomy,torresani2008nonrigid,tomasi1992shape}, and SLAM~\citep{smith1990estimating,castellanos1999spmap}.
Other strategies reconstruct by analysis (e.g., silhouettes~\citep{esteban2004silhouette}) or by synthesis via volume rendering~\citep{kajiya1984ray}, using either implicit representations~\citep{nerf} or explicit ones~\citep{sitzmann2019deepvoxels,liu2020neural}.
Supervised deep learning methods predict voxels~\citep{xie2019pix2vox,wang2021multi}, point clouds~\citep{van2022revealing}, or meshes~\citep{worchel2022multi,wen2019pixel2mesh++}, 
or optimize implicit representations~\citep{liu2024one}, e.g., signed distance functions (SDFs), often with high-quality output but requiring multiple views at inference.
In contrast, we focus on the more restrictive setting of a single RGB image at test time.

\paragraph{Single-view 3D reconstruction} is considerably more difficult.
A large body of work trains reconstruction models with direct 3D supervision, predicting meshes~\citep{xu2019disn,kulkarni2021drdf}, voxels~\citep{girdhar2016learning,wu2017marrnet,popov2020corenet}, point clouds~\citep{fan2017point,mescheder2019occupancy}, or CAD-aligned geometry~\citep{wang2018pixel2mesh,gkioxari2019mesh}.
More recently, large internet corpora of 3D object assets have enabled a line of generative approaches trained using VAE~\citep{kingma2013auto} latent representations~\citep{zhang20233dshape2vecset,xiang2025structured,ren2024xcube}.
A related challenge is recovering the full shape of occluded objects from a partial mask~\citep{xu2024amodal,wu2025amodal3r}; \method handles this through data rather than architecture.
However, existing generative methods are typically evaluated on simplified synthetic single-object benchmarks such as ShapeNet~\citep{chang2015shapenet}, Pix3D~\citep{sun2018pix3d}, or Objaverse~\citep{deitke2023objaverse}.

\paragraph{Layout estimation} extends single-instance estimation to full scenes.
Instead of predicting a single mesh for the entire scene, \method estimates scene layout by using \emph{mask prompts} to indicate which objects to reconstruct, and then estimates both shape and pose for each mask instance.
Approaches can be decomposed into so-called model-based or model-free methods.
Model-based methods decompose the problem into independent parts, first estimating shape (or pulling it from a CAD database) and then posing the objects in the scene~\citep{labbe2022megapose,wen2024foundationpose,maninis2022vid2cad,tyszkiewicz2022raytran,maninis2023cad,rozumnyi2023estimating}.
Alternatively, model-free methods estimate both shape and pose jointly, as in \method~\citep{brazil2023omni3d,huang2025midi,yao2025cast,Ardelean2025Gen3DSR,geng2025one,shi2025sampose}.
However, prior approaches are typically restricted to tabletop robotics, streets, or indoor scenes where objects rest on a supporting surface.
In contrast, our approach estimates both shape and pose for a broad range of object types across diverse scenes.

\paragraph{3D datasets.}
Sourcing 3D annotations is challenging: the modality itself is complex, and the specialized tools required are hard to master. Anecdotally, modeling a 3D mesh from a reference image can take an experienced artist hours (\Cref{section:eval}).
Instead, existing 3D datasets (\eg, ShapeNet~\citep{chang2015shapenet}, Objaverse-XL~\citep{deitke2023objaverse}) primarily consist of single synthetic objects; without paired real-world images, models can only learn from rendered views.
In the real-world domain, existing datasets are small and mostly indoors~\citep{reizenstein2021common,khanna2024habitat,fu20213d,szot2021habitat,pan2023aria}, or procedurally generated~\citep{dai2024automated,infinigen2024indoors,procthor}.
Models trained on such constrained data struggle to generalize.

\paragraph{Post-training.}
While post-training began with a single supervised finetuning stage~\citep{girshick2014rcnn,wei2021flan}, strong pretraining~\citep{brown2020language} made alignment much more data efficient~\citep{hernandez2021scaling}, enabling iterative preference-based alignment like RLHF~\citep{ouyang2022training} and online DPO~\citep{tang2024understanding, rafailov2023direct}.
When post-training must provide a strong steer, self-training methods offer denser supervision--leveraging the model itself to generate increasingly high-quality demonstrations, rather than relying solely on preference signals~\citep{gulcehre2023reinforcedselftrainingrestlanguage,thomas2017exit,dong2023raft,yuan2023rft}.
\method employs self-training to bridge the synthetic$\rightarrow$real domain gap and break the data barrier for 3D perception; most closely resembling RAFT~\citep{dong2023raft}, but also incorporating preference tuning.

\paragraph{Multi-stage pretraining.}
Modern pretraining increasingly employs multiple training stages. Early work on curriculum learning~\citep{bengio2009curriculum} provided a basis for staged data mixing in pretraining, with higher-quality data coming later~\citep{grattafiori2024llama3herdmodels, olmo20252olmo2furious}. \citet{li2023textbooksneediiphi15,abdin2024phi3technicalreporthighly} show that mixing synthetic/web curricula can achieve strong performance at smaller scales.
Increasingly, additional mid-training stages are used for capability injection, such as context extension~\citep{grattafiori2024llama3herdmodels} or coding~\citep{rozière2024codellamaopenfoundation}, and recent work finds that mid-training significantly improves post-training effectiveness~\citep{rlhf2024,wang2025octothinkermidtrainingincentivizesreinforcement}.
\method introduces pretraining and mid-training that can generalize for 3D.

\section{Conclusion}
We share SAM 3D: a new foundation model for full reconstruction of 3D shape, texture, and layout of objects from natural images. SAM 3D's robustness on in-the-wild images, made possible by an innovative data engine and modern training recipe, represents a step change for 3D and an advance towards real-world 3D perception.
With the release of our model, we expect SAM 3D to unlock new capabilities across diverse applications, such as robotics, AR/VR, gaming, film, and interactive media.

\section*{Acknowledgements}
We thank the following individuals for their contributions to this work:

For their contributions to SAM Playground Engineering we thank: Robbie Adkins, Rene de la Fuente, Facundo Figueroa, Alex He, Dex Honsa, Alex Lende, Jonny Li, Peter Park, Don Pinkus, Roman Radle, Phillip Thomas, and Meng Wang.  We thank our excellent XFN team for leadership and support: Kris Kitani, Vivian Lee, Sasha Mitts, George Orlin, Nikhila Ravi, and Andrew Westbury.  Thanks to Helen Klein, Mallika Malhotra, and Azita Shokrpour for support with Legal, Privacy, and Integrity.  We thank Michelle Chan, Kei Koyama, William Ngan, Yael Yungster for all the design support throughout the project.  Thanks to Arpit Kalla for work on model efficiency.  We thank Faye Ma and Kehan Lyu for data engineering support and tooling, and Emmanuel Hernandez, Robert Kuo for pipeline development.  We thank Nan Yang for support with egocentric video data efforts.  Thanks to our two interns Cem Gokmen, Jasmine Shone for their work on 3D and Lea Wilken for feedback on the manuscript. Thanks to our fantastic data operations team: Paris Baptiste, Karen Bergan, Kai Brown, Ida Cheng, Khadijat Durojaiye, Patrick Edwards, Daniella Factor, Eva Galper, Leonna Jones, Zayida Suber, Tatum Turner, Joseph Walker, and Claudette Ward.

\clearpage
\newpage
\bibliographystyle{assets/plainnat}
\bibliography{paper}

\clearpage
\newpage
\beginappendix

\section*{Outline}
The appendix provides additional context to the main paper; it contains additional details about the method and the implementation in \method, as well as ablations.

The structure of the appendix is as follows:
\begin{enumerate}[label=(\roman*)] 
\item \textbf{Data Engine details:} A more detailed description of the data collection used in the \emph{collection step} in \Cref{section:data}.
\item \textbf{Pretraining and Mid-Training Data:} How we collected and filtered the data used for pretraining and mid-training the Geometry and Texture \& Refinement models
\item \textbf{Training details:} Architectural details about MoT and the VAEs. Definitions used for objectives used in each stage. Details on the Geometry and Texture \& Refinement models.
\item  \textbf{Evaluations:} Introducing the details of the new \textbf{SA-3DAO} benchmark, and evaluation protocols of preference tests and quantitative metrics 
\item \textbf{Additional experiments and qualitative examples:} Providing additional analysis and insights into the model's performance.
\item \textbf{Limitations:} An analysis of common failure modes, and future work
\end{enumerate}

\section{Data Annotation Engine Details}
\label{sec:data_eng_details}

\subsection{Stage 1: Image and Object Candidate Sourcing}
\label{sec:data_eng_img_obj_selection}
\paragraph{Image sources.}
To promote generalization across diverse real-world scenes, we expanded our domain coverage by sourcing images from multiple datasets. These include large-scale web-sourced imagery (SA-1B \citep{kirillov2023segment}, MetaCLIP \citep{xu2024metaclip}), video data capturing everyday environments (SA-VI \citep{li2023segmentanyvideo}), egocentric video datasets (Ego4D \citep{grauman2022ego4d}, Ego-Exo4D \citep{grauman2024egoexo4d}, AEA~\citep{lv2024aria}, AEO~\citep{straub24efm}, Nymeria~\citep{ma24eccv}), and domain-specific collections such as food (Food Recognition \citep{bossard2014food101}) and driving scenes (BDD100k \citep{yu2020bdd100k}).

We first filter out images with low resolution, severe blurriness, low contrast, or noticeable artifacts to ensure high-quality visual inputs that are representative of real-world scenarios. Next, we employ visual-language models for object recognition to generate object-level annotations for each image. Images containing only uninformative backgrounds (\eg, ground, sky, ocean) without salient 3D objects are subsequently removed from the dataset.

For each object description, we employ a referral segmentation model to visually ground the object, followed by human annotator verification or refinement of object masks. We discard low-quality masks, masks covering multiple objects, or partial masks that do not capture a distinct object part. This ensures that each retained mask corresponds to a clearly indexable single object instance with sufficient granularity.

\paragraph{2D object selection}
In addition to the objects manually selected and masked by annotators, we also supplement our object mask inputs with segmentation masks sampled from pre-existing datasets.
Besides saving annotation time, this strategy gives us more fine-grained control over the object distribution of the input masks, as object distributions are difficult to enforce on a per-image or per-annotator basis.
To ensure a broad coverage of object categories, we adopt two complementary sampling strategies. First, we construct a 3D-oriented taxonomy by carefully merging and modifying the LVIS~\citep{gupta2019lvis} $1,200$ object categories, emphasizing representations of 3D geometry. For example, different dog breeds are grouped together due to their similar underlying 3D structures, regardless of color, texture, or size. Second, we incorporate human annotator input to identify additional salient objects that may fall outside the taxonomy or are difficult to describe using text alone.

We retain object category labels and continuously monitor the distribution of objects passing through our data engine. To balance throughput and efficiency, we employ a curriculum-inspired sampling strategy, progressing from simple to increasingly complex geometries. Specifically, we begin with rigid objects of simple shapes (\eg, balls, cylinders), transition to more structurally complex objects (\eg, tools, buildings) and ultimately include non-rigid and highly deformable objects (\eg, animals, humans, clothing). The sampling distribution is adaptively adjusted to reflect the evolving dataset composition, with particular emphasis on gradually expanding coverage of long-tail object categories. Through this strategy, we're able to source 850,000 unique object instances from 360,000 images, with annotations covering a wide range of object categories.

\begin{figure*}[t]
  \centering
  \includegraphics[width=1.0\linewidth]{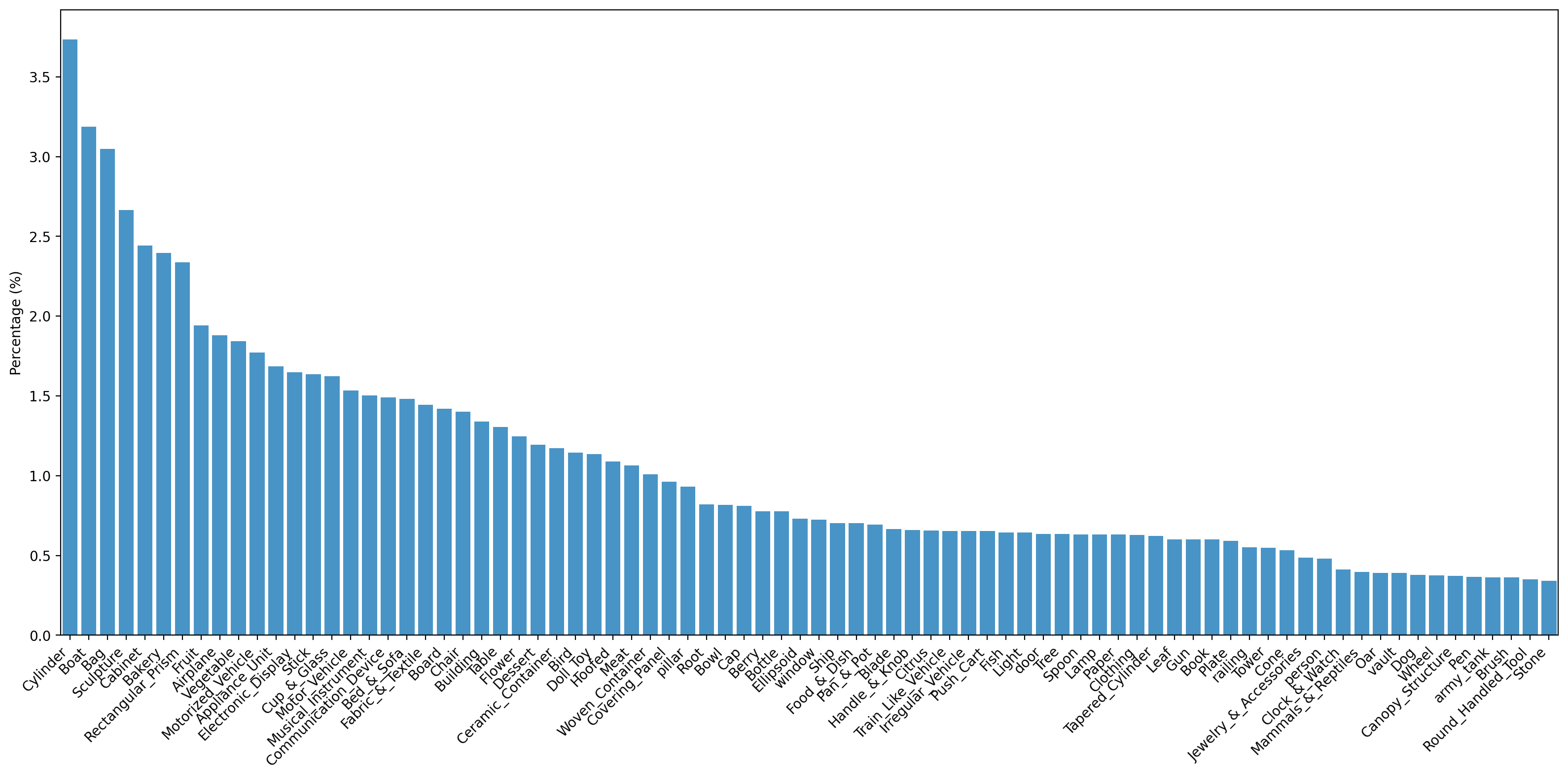}
  \caption{\textbf{Category distribution of \method training data}. The plot above shows the distribution of the top 80 object categories, which includes a long tail.}
  \label{fig:datadist}
\end{figure*}
\paragraph{Texture MITL-3DO.} The MITL-3DO dataset for texture is separate from the dataset for shape and layout, but is collected in a similar fashion. The images are sourced from SA-1B~\citep{kirillov2023segment}, and we additionally sample a dataset of examples with higher aesthetics -- objects with minimal occlusion and high brightness, contrast, colorfulness, sharpness, and aesthetic score -- to seed the model with higher-quality texture annotations. We found the high-aesthetics dataset to further improve human preference rate (see ``AES'' preference win rate in \Cref{fig:ablations_texture}).

\subsection{Stage 2: 3D Model-in-the-Loop Suite}
\label{sec:mitl_suite}
\paragraph{3D shape model suite.}
3D shape generation is beyond the capabilities of the average human annotator, and it is a time-consuming process even for trained specialists (see \Cref{sec:sa3dao}).
Thus, in order to scale shape generation in our annotation pipeline, we instead convert the task to one of verification. 
We achieve this by employing a diverse set of 3D models to generate shape predictions for each object, asking annotators to pick and grade the best of $N$ options.
The sources of 3D shapes in our annotations include the following:
\begin{itemize}
    \item \textbf{Retrieval}: The nearest 3D object is retrieved from a shape object library (pretraining data) using both image- and text-based similarity. For text similarity, we compare visual object descriptions; for image similarity, we compute the distance between CLIP embeddings. While this retrieval approach is nearly guaranteed not to provide an exact 3D reconstruction, it can provide a high quality mesh with matching semantics, particularly when model-generated 3D shapes fail entirely.
    \item \textbf{Text-to-3D generation}: A text-to-3D generative method produces 3D object meshes based on a textual descriptions. This approach can be helpful when image-conditioning is challenging due to clutter or occlusion, but human recognition can still identify the object.
    \item \textbf{Image-to-3D generation}: Image-to-3D methods, including our own \method checkpoint, generate 3D objects in the form of point clouds or meshes, conditioned on the image input. When successful, this tends to produce examples that go beyond semantic matches and better respect the object's physical appearance in the image. However, lack of robustness to occlusion or clutter can negatively impact the results.
\end{itemize}

\paragraph{3D texture model suite.}
For texture generation, we utilize image-to-3D models, multi-view texture generation models, and our own \method checkpoints. All texture candidates are generated using shapes produced by the \method Geometry model, ensuring that texture models have the best chance of success, even in cases of heavy occlusion.

\begin{figure}[t]
  \centering
  \includegraphics[width=0.9\columnwidth]{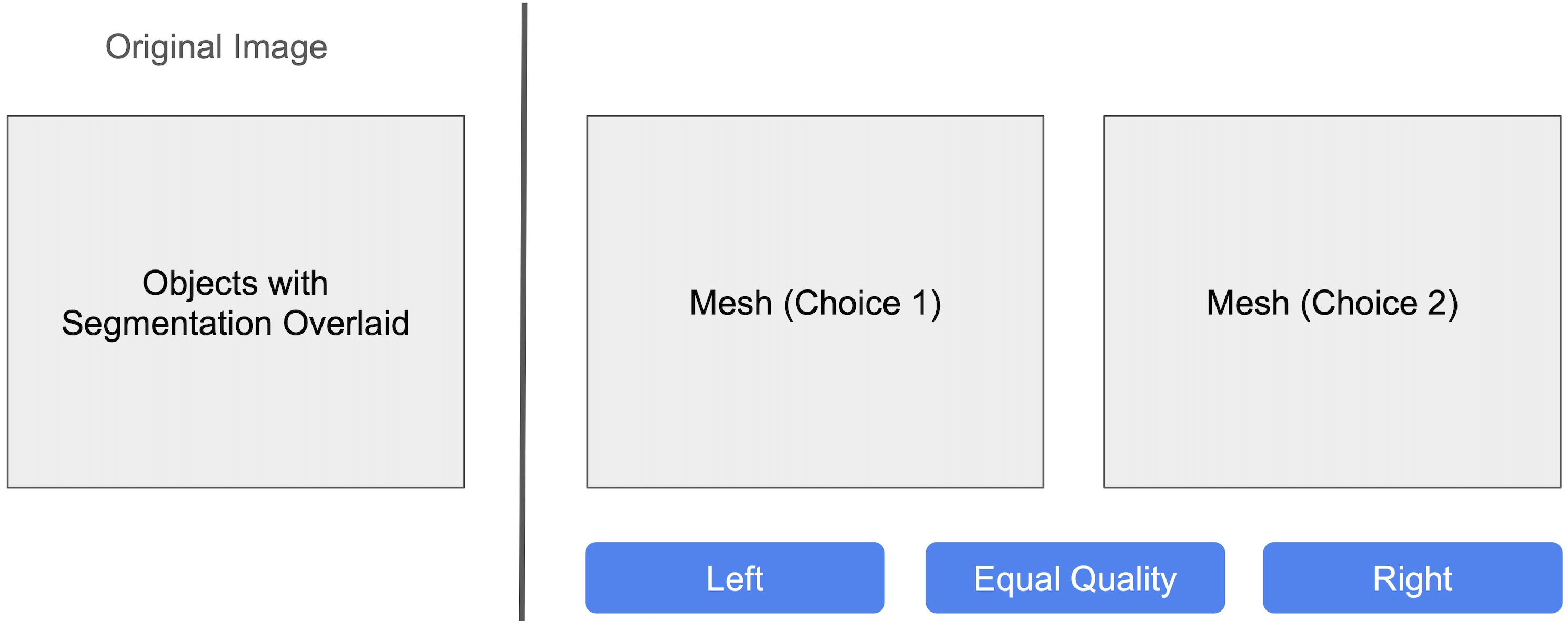}
  \caption{\textbf{Stage 2 UI sketch.} Annotators can only choose between options; they cannot directly edit the meshes or textures.}
  \label{fig:task2_ui}
\end{figure}
\paragraph{Stage 2 Selection procedure.}
The annotators select the best-of-$N$ candidates by making a series of pairwise comparisons (see \Cref{fig:task2_ui}).
For each object, the annotator is initially presented with two candidates to compare and is asked to pick from the following three choices: ``Left'' (is better), ``Right'' (is better), or ``Equal Quality''. 
Because the options are in 3D, we by default automatically rotate the objects on a turntable, but annotators are free to rotate the objects as they wish, or zoom the camera.
After making a selection, the non-selected option is replaced by a new candidate; if ``Equal Quality'', we randomly choose which candidate to keep. 
The selection procedure continues until all candidates have been shown.
We randomize the order in which candidates are presented to the annotator, to prevent biases due to order from affecting the selection process.

After the best candidate is identified in the selection process, annotators are asked to rate the mesh against a predefined quality bar $\alpha$. Examples meeting the bar will become candidates to enter Stage 3 for alignment, while examples under the bar will become negative examples for preference alignment or considered as candidates for manual mesh generation in Stage 2.5.

\subsection{Stage 2.5: 3D Artist Mesh Details}
\label{sec:appx_artist}
When the 3D model-in-the-loop suite fails to generate an acceptable mesh for a particular sample, the aforementioned preference-based annotation approach is unable to provide the data needed to improve the model for such objects. 
To overcome this data distribution blind spot, we work with a team of 3D artists to build meshes for such hard meshes.
Given the high cost of specialized 3D artists, we seek to maximize their value by ensuring each object sent to the 3D artists represents a genuine failure case that cannot be resolved by the data engine alone. To maximize the value of this investment, we develop a refined labeling framework that categorizes failures into common types: \eg, complex geometry, occlusion, transparency, and small object size. We balance sampling across these categories. In addition, we employ clustering techniques over images, 3D latents, and object semantics to deduplicate candidates, ensuring that one or a few representative samples per group suffices for effective coverage in data sampling. 

Additional details on the data collection process for meshes created by 3D artists can be found in \Cref{sec:sa3dao}, which employed a similar mesh creation process by the artists, but with more intentional curation of inputs.

\subsection{Stage 3: 3D Mesh Alignment}
\begin{figure}[t]
  \centering
  \includegraphics[width=0.9\columnwidth]{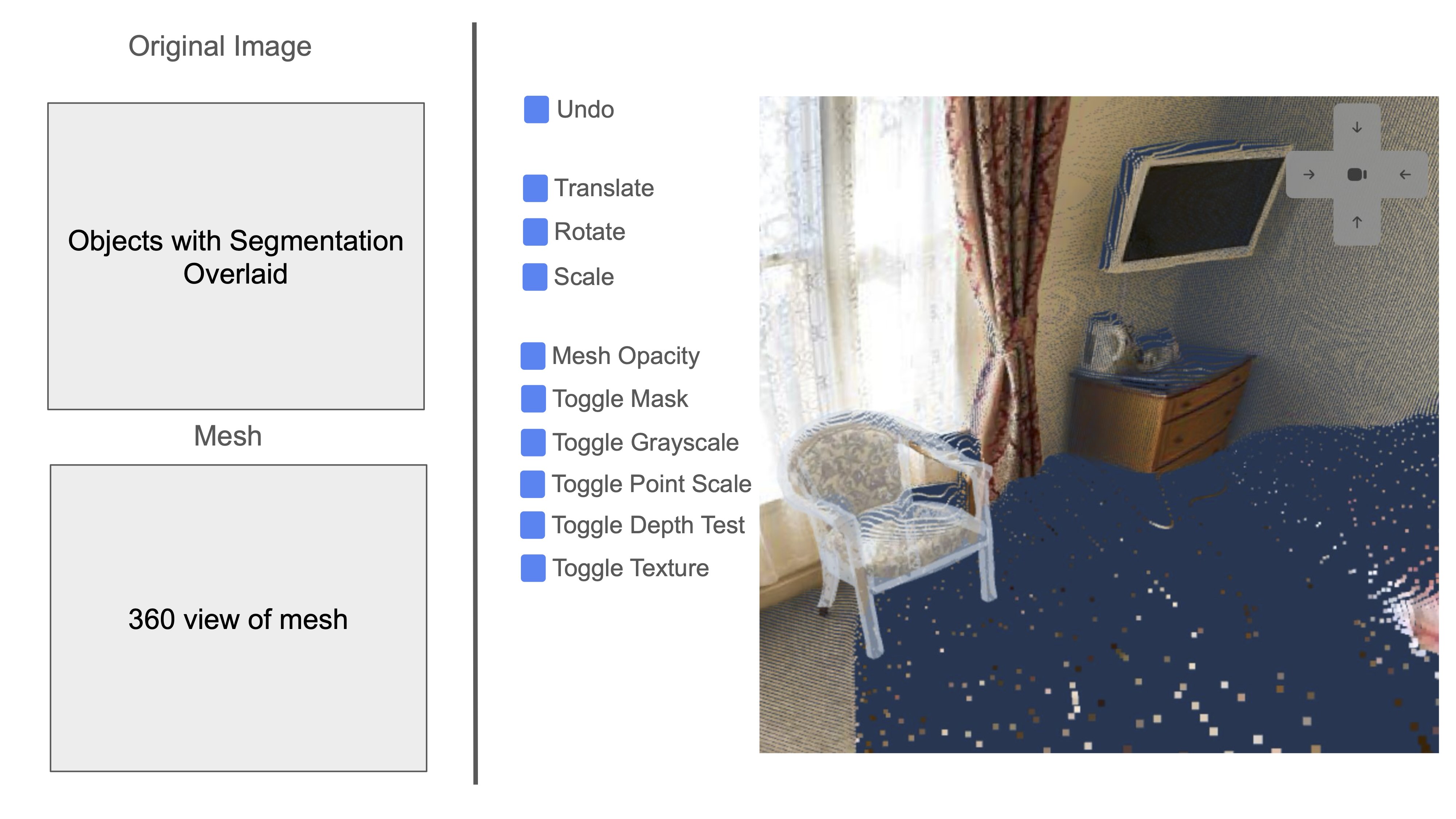}
  \caption{\textbf{Stage 3 UI sketch.} The UI supports annotators in directly placing the object in the 2.5D pointcloud.}
  \label{fig:task3_ui}
\end{figure}

We collect object pose annotations by aligning meshes from prior stages to a scene point cloud derived from the input image.
To make this accessible to generalist annotators, we designed and implemented an annotation tool which allows the annotator to manipulate 3D meshes to align to a 2.5D point cloud pre-computed by an off-the-shelf depth estimator. Annotators can use either keyboard or mouse to rotate, translate, and scale the meshes so that the mesh is accurately anchored to the 2.5D point cloud. We also provide additional functions including (a) mesh visibility toggle, (b) target indicator toggle, (c) point cloud size adjustment, (d) control visibility toggle, (e) undo, (f) camera view reset and pre-defined view, and (g) mesh IOU indicator as shown in \Cref{fig:task3_ui}.

\subsection{Annotation Statistics}
\label{sec:data_engine_statistics}
\begin{itemize}
    \item Stage 1: Annotators on average spend 10 seconds to segment a single interesting object. We utilize SAM \citep{kirillov2023segment} as a tool to assist in segmentation.
    \item Stage 2: Annotators on average spend 80 seconds to select the best candidate shape/texture from 6-10 candidate meshes from variable sources.
    \item Stage 3: Annotators on average spend 150 seconds to anchor and orient the matched 3D shape to the 2.5D point cloud.
    \item Over the lifetime of the project (including development), our MITL data engine yields 3.14 million trainable shapes, 1.23 million samples of layout data, ~100K trainable textures, and over 7 million pairwise preferences.
\end{itemize}

\subsection{Core Alignment Algorithm}
\label{sec:data_engine_online_alg}
\begin{algorithm*}[t]
    \renewcommand{\baselinestretch}{1.1}
    \caption{\method Basic Alignment (Texture, Shape)}
    \label{alg:data_engine}
    \begin{algorithmic}[1]
    \Require Base model $\pi_0$, quality threshold curriculum $\alpha_k$, ensemble size $N$
    \Ensure Aligned model $\pi_K$ 
    \State {\textbf{//} Let $d = (I, M, S, T, R, t, s)$ denote a demonstration (i.e., a training sample)}
    \For{$k = 1$ to $K$}
        \State {\textbf{// Collection Step:} Generate demonstrations via expert policy}
        \State Initialize $\mathcal{C}_k \leftarrow \emptyset$ \Comment{The dataset collected during iteration $k$}
        \For{$(I, M) \sim p(\mathbf{I},\mathbf{M})$}
            \State $\tilde{\pi}_k \leftarrow \text{Amplify}(\pi_{k-1})$ \Comment{Amplify current policy via model ensemble and best-of-$N$ search}
            \State Sample $\{d_i\}_{i=1}^N \sim \tilde{\pi}_k(I, M)$ \Comment{Generate $N$ candidate demonstrations from expert policy}
            \State $d^*, r \leftarrow \text{HumanRank}(\{d_i\}_{i=1}^N)$ \Comment{Humans select best candidate via pairwise comparisons}
            \State $\mathcal{R} \leftarrow \{d_i : i \neq \arg\max\}$ \Comment{Store rejected candidates for preference learning}
            \State $\mathcal{C}_k \leftarrow \mathcal{C}_k \cup \{(d^*, r, \mathcal{R})\}$ \Comment{Collect chosen demonstration with rating and rejections}
        \EndFor
        \State {\textbf{// Update Step:} Train on aggregated high-quality demonstrations and preferences}
        \State $\mathcal{C} \leftarrow \{(d^+,\mathcal{R}) : (d^+, r, \mathcal{R}) \in \bigcup_{i=1}^k \mathcal{C}_i, r \geq \alpha_k \}$ \Comment{Aggregate and filter by quality}
        \State $\mathcal{D} \leftarrow \{(d^+, d^-) : (d^+, \mathcal{R}) \in \mathcal{C}, d^- \in \mathcal{R}\}$ \Comment{Create preference pairs for DPO training}
        \State $\pi_k^{\text{SFT}} \leftarrow \arg\min_\pi \mathbb{E}_{(d^+, d^-) \sim \mathcal{D}} [\mathcal{L}_{\text{CFM}}(\pi; d^+)]$ \Comment{Supervised finetuning}
        \State $\pi_k \leftarrow \arg\min_\pi \mathbb{E}_{(d^+, d^-) \sim \mathcal{D}} [\mathcal{L}_{\text{DPO}}(\pi, \pi_k^{\text{SFT}}; d^+, d^-)]$ \Comment{Align with preferences}
    \EndFor
    \State \Return $\pi_K$
    \end{algorithmic}
\end{algorithm*}

\subsubsection{Basic Algorithm}
\Cref{alg:data_engine} shows the core alignment algorithm, used for all texture annotations and most shape annotations (MITL-3DO). During each collection step, we generate a set of predictions from the current model, and ask annotators to rank and verify these predictions. Generalist annotators can only choose between model outputs and accept/reject; they cannot edit.
We maximize the probability of a successful annotation at each iteration by ensembling multiple models and combining multiple models with human preferences into an \emph{expert} annotator.

The learning efficiency of the alignment, or the ``speed of the data flywheel'', is controlled by two factors:
\begin{itemize}
    \item \textbf{Amplification factor}: The size of the performance gap between current model and the expert annotations at each iteration
    \item \textbf{Stepwise efficiency}: How closely the new model approximates the previous expert from the previous iteration
\end{itemize}
The former induces an upper bound on the new policy's performance at each iteration, while the latter describes how closely we approach that upper bound -- similar to Expert Iteration~\citep{thomas2017exit}.

\subsubsection{Training Intuition}

Our goal in post-training (see \Cref{alg:data_engine}) is to align the model to match human preference on the distribution of \emph{all} possible real-world objects.

The core algorithm in our data engine generates samples by asking humans to select viable samples from a set of candidate generations. Challenging inputs often result in no viable candidate generations and thus never get selected by humans. However, at any current time our model is usually good on some parts of the data distribution, but not on other parts, as shown in the cartoon below:

\begin{figure}[t]
    \centering
    \includegraphics[width=\columnwidth]{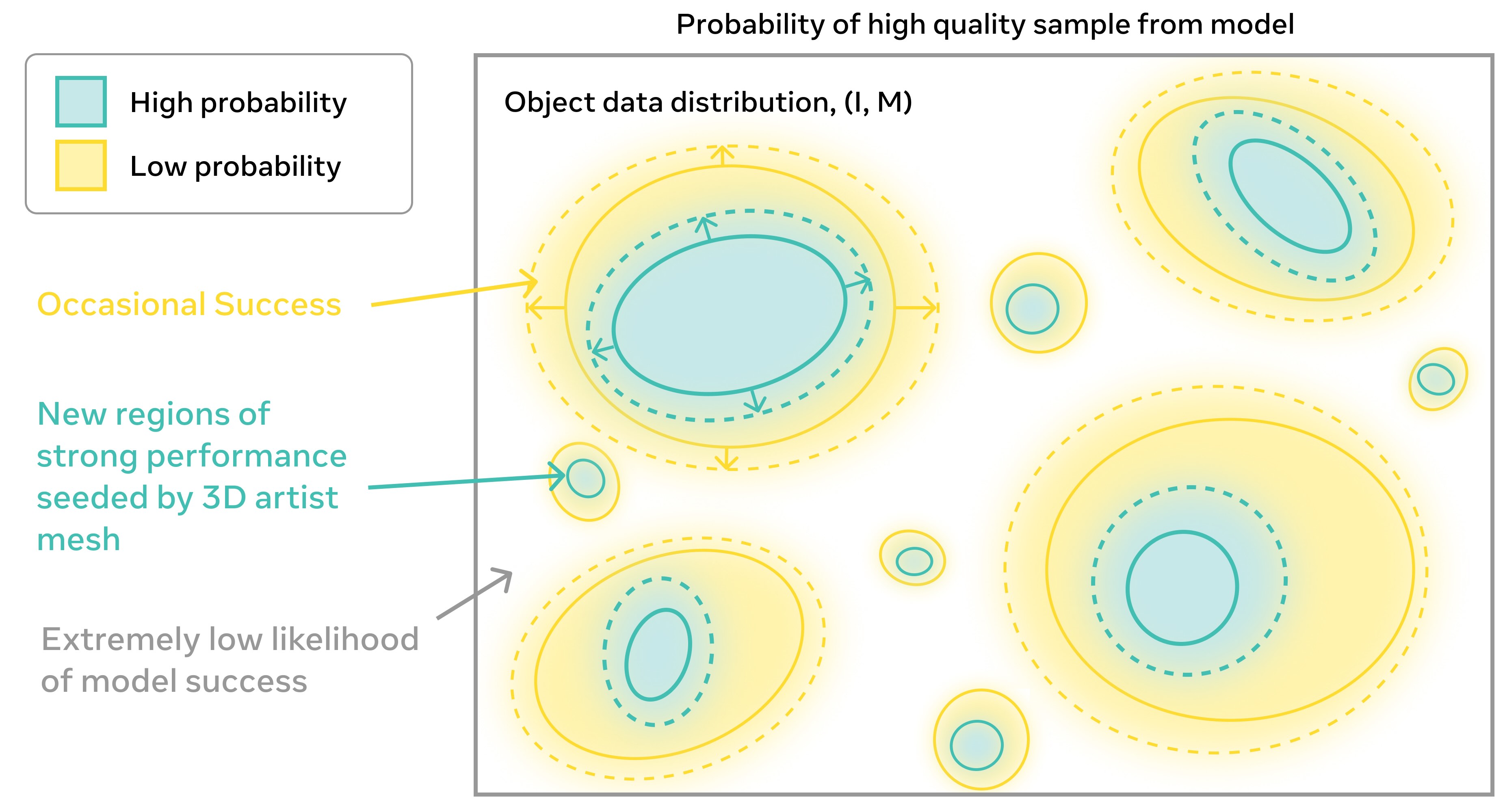}
    \vspace{-3mm}
    \caption{
        \textbf{Simplified cartoon depiction of data engine improvement.} 
        The diagram depicts model sample quality (color) across the real-world distribution of images and masks.
        During training, the model begins by doing well \textcolor{teal}{(teal)} on common categories and simple objects (chairs, bottles, signs, cars). Our goal is to both improve accuracy and robustness on these easy examples \textcolor{teal}{(\textbf{teal})}, and then push the model to improve performance on less common objects  \textcolor{readableyellow}{(\textbf{yellow})} in the tail of the probability distribution. While the amplification stage of MITL generally leads to the slow expansion of existing regions of success, using 3D artists to create data for the hardest samples allows us to shortcut the process by ``seeding'' new regions of the data distribution, which may have taken us longer to reach through MITL verification alone.
    }
    \label{fig:pushing_into_tail}
\end{figure}

The intuition behind the data engine in \method is that these green islands of reliably good performance correspond to high-density parts of the training data~\citep{oneill2024openx,nvidia2025gr00t}, and the approach in \method is that we want to push out from these islands of reliably good generations into the ``tail'' of the distribution, demarcated by yellow and white background in the cartoon above. The yellow parts of the distribution are challenging for the model, but near enough to the blue islands, that we can \emph{occasionally} generate satisfactory annotations, but it requires humans to go through many samples.

This can create a chicken-and-egg problem where, for the model to become good, it must already be capable of produce a good generation; at least some of the time. For examples that are so challenging that the probability of success is extremely low (white), the model has no hope and we ask human 3D artists (\Cref{sec:appx_artist}) to provide supervision in this part of the data distribution, in order to seed new islands. 

\subsection{Increasing Amplification Factor with Search}
\label{sec:test_time_search}

\subsubsection{Best-of-$N$ Search with Reward Models}
\begin{figure}[t]
    \centering \includegraphics[width=\linewidth]{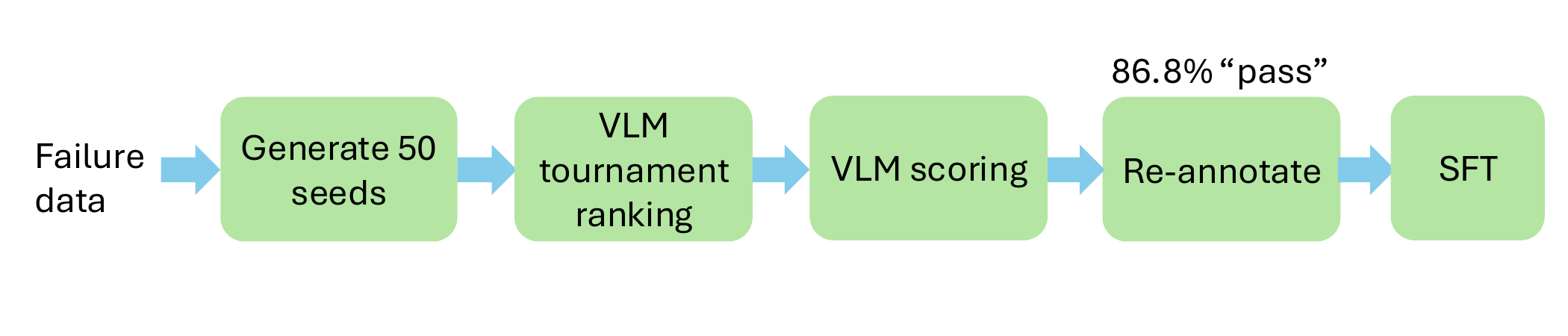}
    \caption{\textbf{Reward model data recovery pipeline}. The diagram shows how we use reward mdoels to increase $N$ in best-of-$N$ search to improve the chance of a successful annotation on challenging tail inputs. We use both a VLM and also DPO implicit reward as reward models.}
    \label{fig:vlm_search}
\end{figure}

Qualitatively, however, we observe that re-visiting some (yellow) inputs with a large number of seeds, our model can sometimes still yield a few good generations (\eg, food can take around $\sim50$ seeds to reliably generate a successful mesh in~\Cref{table:vlm_search}). This suggests that increasing $N$ in the best-of-$N$ rejection sampling can potentially allow us to obtain annotations for challenging inputs, which would be difficult to source otherwise. Doing so would allow us to rapidly ``push into the tail'', increasing the convergence speed of the alignment algorithm in~\Cref{alg:data_engine}. However, the primary impediment to increasing $N$ is that, at some point, there are too many choices for a human to compare. This linearly scales the annotation time of preference data collection, and the selections themselves become noisier and more random due to choice overload~\citep{diehl2010expectations}.

To address this challenge, we explored using learned reward models to perform a first pass in order to surface a smaller number of candidates for humans to then choose between. 
\Cref{fig:vlm_search} shows a pipeline to perform reward-ranked best-of-$N$ search that increases the yield of successful annotations on challenging inputs. We first run $50$ generations with different initial noise, and use the reward model to perform tournament-style ranking, and then pass the winning candidate to human annotators for ranking and verification (as in Stage 2).  

We find that this approach indeed helps to recover some of this otherwise difficult data. For example, by scaling the best-of-$N$ from $N=2$ to $N=50$ to recover samples that were originally discarded, improving the yield from $0\%$ (since these were originally failures) to $86.8\%$. In particular, we observe significant increase in the proportion of successful annotations coming from challenging categories. The \emph{food} category improves $9 \times$ from $4\%$ in the original annotated distribution to $36\%$. We show the experiments with resulting model performance, as well as ablations using VLMs instead of DPO implicit reward models, in~\Cref{sec:best_of_n_ablation}.

\subsection{Relationship to Self-Training Approaches}
The data engine in \method can alternatively be viewed as an online alignment algorithm similar to RLHF~\citep{ouyang2022training} or related self-training methods. Under this interpretation, the generative model $q$ is a policy and the data collection step is a policy evaluation; collecting demonstrations $D^{+}$ and preferences $D^{+}$/$D^{-}$ through the interaction with the environment (annotators). The model improvement step simply updates the current policy using both finetuning and DPO.

This reframing helps make the relationship to existing work more clear. The most similar learning algorithm to our data engine is Expert Iteration (ExIt)~\citep{thomas2017exit}. As in ExIt, each iteration starts with a current policy, that we amplify using additional information into an expert policy, and we use this expert policy to generate supervision for imitation learning. Unlike ExIt, which uses purely imitation learning, we use humans-as-verifiers to select which samples to train on, and we make use of additional preference signals as reward signal (\Cref{sec:dpo}).
However, there are also notable differences in type of supervision that can be used and the amplification steps. Our expert policy amplification step uses a model ensemble instead of only tree search with a value function, and we use preferences in the update step to better align to the task.

\Cref{alg:data_engine} uses reward ranking, similar to RAFT~\citep{dong2023raft} and RFT~\citep{yuan2023rft}, although the alignment algorithm in \method adds explicit expert policies/ensembles and leverages preference supervision.

\section{Pretraining and Mid-Training Data Details}

\subsection{Iso-3DO Data Filtering}
\label{sec:iso_preprocessing}
For the Iso-3DO data used for pretraining, the quality of the 3D meshes can vary substantially, and not all samples exhibit high-fidelity geometry. 
Such examples can ultimately prove harmful to model pretraining, even at scale.
One way to filter data is by an aesthetic score filter, as employed by~\citet{xiang2025structured}, which primarily emphasizes visual and textural appeal. We employ a similar filter process for the Texture \& Refinement model. 

However, this filter does not necessarily capture the geometric quality of a training data. Therefore, we develop a rule-based filtering strategy on shape to curate the pretraining data for the Geometry model, removing data with following characteristics:
\begin{itemize}
    \item \textbf{Overly simplistic geometry}, characterized by extremely small volumes (\eg, near-degenerate point-like structures) or minimal normal direction variation (\eg, flat, sheet-like surfaces).
    \item \textbf{Structural outliers}, which includes meshes containing spatial outliers: isolated points or fragments that deviate significantly from the primary 3D structure.
\end{itemize}

\subsection{Render-and-Paste Data Pipeline}
\label{app:mid_training_data}

We define the Render-Paste approach as follows:
Given a natural image and an object instance defined by its mask segment, we replace the object in the image with a synthetic 3D object drawn from the same synthetic sources used in Iso-3DO. 
The size and position of the 3D object are determined using the 2D object mask together with a pointmap produced by a single-image depth estimator, which also guides the object's visibility and occlusion to obtain a natural appearance in the final rendering.
By nature of starting from a synthetic 3D object first, the resulting data (which we refer to as \emph{RP-3DO}) has excellent 3D ground truth precision and pixel-alignment compared to subsequent data sources in our training pipeline, which much try to reconstruct 3D from partial 2D information as part of the data annotation process.

In the following sections, we introduce three variants of Render-Paste that differ along two axes: pose information and semantic relevance. 
\begin{itemize}
    \item \Cref{sec:flying_occlusion}: \textit{Flying Occlusions} \textbf{(FO)} inserts randomly oriented synthetic objects without pose information, resulting in pose-unaware but semantically loose composites.
    \item \Cref{sec:eitr}: \textit{Object Swap -- Random} \textbf{(OS-R)} determines object scale and translation from masks and pointmaps, while using a random rotation and object. Beyond simple replacement, the incorporation of depth ordering provides meaningful visual cues for object size and spatial placement, yielding pose-aware but not fully aligned insertions with moderate semantic relevance, higher than in the Flying Occlusions setting.
    \item \Cref{sec:render_paste}: \textit{Object Swap -- Annotated} \textbf{(OS-A)} replaces the original object using the annotator-provided ground-truth scale, translation, and rotation, producing fully pose-aligned and semantically matched renderings.
\end{itemize}

\subsubsection{Flying Occlusions (FO)}
\label{sec:flying_occlusion}

The aim of this dataset is to build invariance to occlusion and size variations that commonly occur in real-world scenarios---and to enable the model to leverage full image context instead of only object-centered crops---we construct a dataset of natural images with blended synthetic 3D objects.
Inspired by Flying Chairs~\citep{dosovitskiy2015flownet} and FlyingThings3D~\citep{mayer2016large}, we name our first variant Flying Occlusions, reflecting its use of freely inserted synthetic objects.

Each training example consists of a natural image onto which we composite two 
rendered 3D objects: an \emph{occluder} and an \emph{occludee}. For each pair, 
we also compute the final visible mask of the occludee after occlusion.
To generate each training sample, we randomly pair a selected object with an occluder object. 
Given the mask of the selected object $M_{\mathrm{obj}}$ and the mask of the random occluder $M_{\mathrm{occluder}}$, 
each corresponding to the full mask of the respective object, 
the visible mask is defined as 
$M_{\mathrm{vis}} = M_{\mathrm{obj}} \odot (1 - M_{\mathrm{occluder}})$,
where $\odot$ denotes the element-wise (Hadamard) product.
To ensure a reasonable degree of occlusion, we enforce 
$0.1 \leq |M_{\mathrm{vis}}| / |M_{\mathrm{obj}}| \leq 0.9$.
In addition, samples with insufficient visibility are filtered out by requiring 
$|M_{\mathrm{vis}}| / |I| \geq 0.2\%$,
where $|I|$ is the total number of pixels in the image.
Here, $|M|$ denotes the sum of all elements in $M$ (\ie, the total number of pixels with value $1$ if $M$ is a binary mask).

Finally, to prevent the model from always predicting the occluded object, in one third of the samples, we treat the selected mesh as the occluder. In these cases, the mask of the selected mesh is complete. In total, we have 55.1M sample with 2.87M unique meshes and 11.17M unique images.

\subsubsection{Object Swap -- Random (OS-R)}
\label{sec:eitr}

To enhance robustness to variations in object location and scale, we propose Object Swap -- Random (OS-R), a depth-aware render-paste strategy that replaces an object in a natural image with a randomly selected synthetic mesh.

Given a natural image $I$, mask $M$, and random object mesh $S$, we synthesize a new training tuple $(I', M_{\mathrm{vis}}, S, R, t, s)$. We first predict the 2.5D scene pointmap and identify the 3D centroid and bounding box of the target object. The original object is removed via inpainting, and we then insert a random synthetic mesh $S$ at the computed centroid $t$ with a random 3D rotation $R$. The mesh scale $s$ is determined by fitting the mesh to the original object's 3D bounding box.

We complete the process by re-rendering the new image $I'$ with a z-buffer check. We render the new mesh into the inpainted image such that only pixels not occluded by existing scene geometry are visible, forming the visible mask $M_{\mathrm{vis}}$. We filter samples with insufficient visibility ($<20\%$ visibility) and update the pointmap $P$ by projecting the unoccluded surface points of the new mesh, using $M_{\mathrm{vis}}$.

To ensure the dataset provides sufficient visual cues for estimating translation and scale, we use heuristics to replace only objects that are partially occluded or supported along the bottom, which provides depth ordering and T-junction cues, respectively. We verify these cues by trying to find occlusion boundaries: we sample points on opposite sides of the mask border, and if the outer pixel is significantly closer to the camera than the inner pixel, we consider this part of the boundary occluded. A sample is retained if it meets one of two conditions: (1) \textit{physical support}, where the background is closer to the camera along the bottom 10\% of the object (indicating it rests on a surface), or (2) \textit{partial occlusion}, where foreground elements occlude at least 10\% of the total object perimeter. This process yields 5.95M training samples composed of 2.38M unique meshes and 1.20M unique images.

\subsubsection{Object Swap -- Annotated (OS-A)}
\label{sec:render_paste}
In addition to the \emph{Object Swap -- Random} variant, we construct a complementary render-and-paste setting, which we refer to as \emph{Object Swap -- Annotated} (OS-A), which performs an in-place replacement of a real image with a rendered human-annotated object. The motivation for this dataset is to enable Texture \& Refinement training that faithfully preserves pixel-aligned correspondence between the rendered mesh and the visual appearance of the target object in the image.

This approach closely follows the \emph{OS-R} pipeline, with key distinctions arising from the use of human-annotated data in MITL-3DO. Specifically, each training sample is generated using an image from a curated MITL-3DO subset, where the initial object mask, selected mesh $S$, object placement (translation $t$, rotation $R$, and scale $s$), and target pose are all sourced from human annotations provided in the MITL-3DO dataset. The selected mesh for each object is chosen by annotators as the best available match to the object's appearance in the image. During rendering, lighting conditions used for rendering are carefully matched to those in the training data preparation, ensuring consistent brightness and appearance across the dataset.
We used a subset of the MITL-3DO shape-preference annotations, yielding $0.4$ million training samples from this render-paste process.

\subsection{Lighting for Texture Data} 
For Iso-3DO and RP-3DO (FO), we randomize the lighting (\ie, direction and intensity) applied on the input images, and use ambient lighting when rendering views used to computing the target latents. Qualitatively, such data processing encourages the model to predict ``de-lighted'' textures without baking in strong directional shading or specular highlights from the input render. We verify that this is preferred by humans through preference tests (see ``Lighting Aug'' preference rate in \Cref{fig:ablations_texture}).

\section{Details on Model Training}

The following sections outline the details of the Geometry and Texture \& Refinement models, including architecture, training objective, and training hyperparameters. 
\subsection{Architecture Details on Geometry Model}
\label{sec:fm_details}

We employ a latent flow matching model. For shape, it denoises the $64^3$ voxels in the latent space of a coarser $16^3 \times 8$ representation, following \citet{xiang2025structured}. For layout, we perform denoising directly in the parameter space $(R, t, s)$, as their dimensionality is small. Additionally, we introduce modality-specific input and output projection layers to map both the shape and layout parameters into a shared feature space of dimension $1024$, and subsequently project them back to their respective parameter spaces. This results in a total of $4096$ tokens for the shape and 1 token for $R, t, s$, respectively, as input to the Mixture of Transformers (MoT). The MoT architecture comprises two transformers: one dedicated to the shape tokens, and a second whose parameters are shared for the layout parameters $(R, t, s)$, as shown in~\Cref{fig:method}.

The MoT design allows independently training of some modalities while maintaining performance on others (\eg, fine-tune shape or layout only), thanks to the structured attention mask illustrated in~\Cref{fig:method}. This proves helpful when training on datasets that contain labels for only one modality (\eg shape-only), and when freezing shape capabilities and finetuning just for layout. At the same time, MoT still allows for information sharing during the forward pass, through the joint self-attention layers for cross-modal interaction. This shared context is critical for self-consistency: notably, rotation is only meaningful when anchored to the predicted shape.

\subsection{Pretraining \& SFT Objective: Conditional Rectified Flow Matching}
\label{sec:flow_matching_training_objective}
SAM 3D is trained to jointly generate multiple 3D modalities using rectified conditional flow matching~\citep{liu2022flow}. Given an input image $I$ and mask $M$, the Geometry model optimizes the following multi-modal flow matching objective:
\begin{equation}
\mathcal{L}_{\text{CFM}} = \sum_{m \in \mathcal{M}} \lambda_m \underset{\raisebox{-1.3ex}{$\scriptstyle \tau, \mathbf{x}_0^m$}}{\text{\Large$\mathbb{E}$}} \left[ \|\mathbf{v}^m - \mathbf{v}_\theta^m(\mathbf{x}_\tau^m, c, \tau)\|^2 \right]
\label{eq:fm}
\end{equation}
where $\mathcal{M} = \{S, R, t, s\}$ denotes the set of prediction modalities (shape, rotation, translation, scale),  $c=(I,M)$ contains the conditioning modalities (image, mask), and $\mathbf{v}_\theta^m$ is the learned velocity field for modality $m$ at the partially noised state, $\mathbf{x}_\tau^m$.

We want to learn to generate clean states $\{\mathbf{x}_1^m\}_{m \in \mathcal{M}} \sim p(\mathcal{M}|c)$, and during training these are the ground-truth 3D annotations for each modality. Then, the target probability path at time $\tau \in [0,1]$ is a linear interpolation $\mathbf{x}_\tau^m = \tau\mathbf{x}_1^m + (1-\tau)\mathbf{x}_0^m$ between the target state $\mathbf{x}_1^m $ and intial noise state $\mathbf{x}_0^m \sim \mathcal{N}(0, \mathbf{I})$. As a result, the target velocity field is the gradient of this linear interpolation $\mathbf{v}^m = \dot{\mathbf{x}}_\tau^m =(\mathbf{x}_1^m{-}\mathbf{x}_0^m)$. $\lambda_m$ is simply a per-modality weighting coefficient. 

The Texture \& Refinement model optimizes analogous flow-matching objectives using SLAT features. We train both models using AdamW (without weight decay), and training hyperparameters such as sampling and learning rate schedules, EMA weights are in~\Cref{sec:texture_hyperparameters}.

We find it noteworthy that flow matching objective is sufficient for SAM 3D to learn the task of 3D reconstruction, without explicitly enforcing geometric constraints through loss objectives.
Interestingly, we observed empirically that while such losses can be helpful in low-data and low-compute regimes, their value declines as data and compute increase, suggesting these geometric priors can be learned implicitly with scale. 

\subsection{Preference Alignment Objective: DPO}
\label{sec:dpo_training_objective}

For preference alignment in \Cref{sec:dpo}, we follow Diffusion-DPO~\citep{wallace2024diffusion} and adapt to flow matching as follows: given the same input image and mask $c$, we sample a pair of 3D output $(x_0^w,x_0^l)$ based on human preference, where $x_0^w$ is the preferred option and $x_0^l$ is the less preferred. Our training objective is:

\begin{align}
\mathcal{L}_\mathrm{DPO} &= - \mathbb{E}_{\begin{array}{rl}
I &\sim~\mathcal{I}, \\
(x_0^w,x_0^l) &\sim~\mathcal{X}_I^2 \\
\tau &\sim~\mathcal{U}(0,T) \\
x_\tau^w &\sim~q(x_\tau^w\mid x_0^w) \\
x_\tau^l &\sim~q(x_\tau^l\mid x_0^l)
\end{array}} 
\big[\operatorname{\log \sigma}\big(-\beta T w(\tau) \cdot \Delta\big) \big] \\
\\ \notag
\text{where} \quad \Delta &= 
\|\mathbf{v}^w - \mathbf{v}_\theta(x_\tau^w,c,\tau)\|_2^2
- \|\mathbf{v}^w - \mathbf{v}_{\mathrm{ref}}(x_\tau^w,c,\tau)\|_2^2 \notag \\
&\quad - \big(
\|\mathbf{v}^l - \mathbf{v}_\theta(x_\tau^l,c,\tau)\|_2^2
- \|\mathbf{v}^l - \mathbf{v}_{\mathrm{ref}}(x_\tau^l,c,\tau)\|_2^2
\big)\notag
\label{eq:dpo}
\end{align}

where $\mathbf{v}^w$ and $\mathbf{v}^l$ are the target flow-matching velocities
for $x_\tau^w$ and $x_\tau^l$, and $\mathbf{v}_\theta$, $\mathbf{v}_{\mathrm{ref}}$ are the
learned and frozen reference velocity fields, respectively.

\paragraph{Implementation details.}
We apply DPO on shape prediction in the Geometry model and the predictions of the Texture \& Refinement model. We use the preference data collected in Stage 2, where we remove the negatives from non-\method generations (\eg retrieval-based methods or multi-view diffusion texture generations), since they are out of the distribution for \method.

\subsection{Model Distillation Objective: Shortcut Models}
\label{sec:inference}

For applications needing online 3D perception capabilities (\eg robotics), model inference time is an essential consideration. In diffusion and flow matching models, the most straightforward way to improve inference speed is by reducing the number of function evaluations (NFE). However, na\"ively decreasing the number of steps can significantly degrade performance.
Instead, we employ flow matching distillation techniques to reduce the number of inference steps while minimizing impact to quality. Specifically, we adopt the diffusion shortcut formulation from~\citet{frans2024one}, which offers several advantages over previous consistency distillation approaches: (1) it is simple, avoiding multi-stage training and instability; and (2) the model supports two modes, allowing seamless switching back to the original flow matching inference, so a single model can serve both purposes. Unlike the original formulation, we do not train shortcut models from scratch. Instead, we initialize from fully trained checkpoints and further finetune them with the shortcut objective. 

\begin{align}
\mathcal{L}_S(\theta)
&=
\mathbb{E}_{
    \substack{
        x_0\sim\mathcal{N}(0,I),\\
        x_1\sim p(x),\\
        \tau,d\sim p(\tau,d)
    }
}
\Big[\qquad
    \underbrace{
    \left\|
        \mathbf{v} - \mathbf{v}_\theta(x_\tau,\, c,\, \tau,\, d{=}0)
    \right\|^2}_{\text{Flow-Matching}}
\qquad
+\qquad
    \underbrace{
    \left\|
        \mathbf{v}_{\mathrm{consistency}} - \mathbf{v}_\theta(x_\tau,\, c,\, \tau,\, 2d) 
    \right\|^2}_{\text{Self-Consistency}}
\qquad  \Big].
\end{align}
\noindent where:
\begin{itemize}
    \item $x_0 \sim \mathcal{N}(0,I)$:
    a Gaussian noise sample drawn from the standard normal distribution.

    \item $x_1 \sim p(x)$:
    a real data sample from the data distribution.

    \item $x_\tau$:
    an interpolated sample between $x_0$ and $x_1$ at time step $\tau$.
    (Defined earlier in the paper through the diffusion / flow matching path.)

    \item $\tau$:
    the diffusion time (or noise level) at which the model predicts a local
    velocity or update step.

    \item $d$:
    the step size specifying how large a step the shortcut model
    should predict.
    $d=0$ corresponds to flow-matching,
    $d > 0$ corresponds to consistency training.

    \item $c$:
    conditioning tokens.

    \item $p(\tau,d)$:
    the joint sampling distribution over diffusion times and step sizes
    used during training.

    \item $\mathbf{v}_\theta(x_\tau,\, c,\, \tau,\, d)$:
    the shortcut model parameterized by $\theta$, taking as input the
    current sample $x_\tau$, conditioning $c$, time $\tau$, and desired step
    size $d$.

    \item $\mathbf{v}$:
    the empirical instantaneous velocity of the data flow
    used as the target for the flow-matching objective
    (corresponds to $d=0$).

    \item $\mathbf{v}_{\mathrm{consistency}}$:
    the self-consistency target,
    constructed by composing two steps of size $d$
    to form a reference for a single jump of size $2d$. \Cref{alg:consistency} describes how to construct $\mathbf{v}_{\mathrm{consistency}}$.
\end{itemize}

\begin{algorithm*}[t]
\caption{Consistency Target Construction with CFG Guidance for Shortcut Model Distillation}
\label{alg:consistency}
\begin{algorithmic}[1]
\Require Current state $x_\tau$, conditioning $c$, step size $d$, CFG weight $w_{\mathrm{CFG}}$, teacher model $\mathbf{v}_\theta$
\Ensure Consistency target $\mathbf{v}_{\mathrm{consistency}}$
\State {\textbf{//} First shortcut step with CFG guidance}
\State $\mathbf{v}_\tau \leftarrow \mathbf{v}_\theta(x_\tau,\varnothing,\tau,d) + w_{\mathrm{CFG}} \big( \mathbf{v}_\theta(x_\tau,c,\tau,d) - \mathbf{v}_\theta(x_\tau,\varnothing,\tau,d) \big)$ \Comment{Apply CFG to get guided velocity}
\State $\tilde{x}_{\tau+d} \leftarrow x_\tau + d \cdot \mathbf{v}_\tau$ \Comment{Take first step of size $d$}
\State \textbf{//}{ Second shortcut step with CFG guidance}
\State $\mathbf{v}_{\tau+d} \leftarrow \mathbf{v}_\theta(\tilde{x}_{\tau+d},\varnothing,\tau+d,d) + w_{\mathrm{CFG}} \big( \mathbf{v}_\theta(\tilde{x}_{\tau+d},c,\tau+d,d) - \mathbf{v}_\theta(\tilde{x}_{\tau+d},\varnothing,\tau+d,d) \big)$ \Comment{Apply CFG at new state}
\State $\tilde{x}_{\tau+2d} \leftarrow \tilde{x}_{\tau+d} + d \cdot \mathbf{v}_{\tau+d}$ \Comment{Take second step of size $d$}
\State \textbf{//}{ Compute consistency target from two-step trajectory}
\State $\mathbf{v}_{\mathrm{consistency}} \leftarrow \mathrm{stopgrad}\Big( \frac{\tilde{x}_{\tau+2d} - x_\tau}{2d} \Big)$ \Comment{Average velocity over combined $2d$ step}
\State \Return $\mathbf{v}_{\mathrm{consistency}}$
\end{algorithmic}
\end{algorithm*}

We also distill CFG into the shortcut mode by using a fixed CFG strength of $w_{\mathrm{CFG}}=2$ for Stage 1 and CFG strength of $w_{\mathrm{CFG}}=1$ for Stage 2. The final model is fine-tuned for approximately $4$K iterations using the same objective as in~\citet{frans2024one}: $75\%$ flow matching and $25\%$ shortcut. When shortcut mode is disabled, the model behaves identically to the original flow matching model. We initialize the step size embedder by setting the weights and bias of its final linear layer to zero, since, unlike the other parameters that have already been trained, these are new parameters introduced at the distillation stage. \Cref{fig:distillation} shows quantitative results and \Cref{fig:distillation_steps_vis} shows examples.

\subsection{Texture \& Refinement Training Details}
\label{app:texture_training}

We train the Texture \& Refinement model following a multi-stage training paradigm analogous to that of the Geometry model (described in \Cref{sec:training}). Below we provide implementation details for the texture training stages.

\paragraph{VAE Training.} Learning the inverse problem of image to texture map requires a strong alignment in training data between them. However, in previous work~\citep{xiang2025structured}, renderings of meshes create artifacts like reflections and shadows; objects consistently having extremely dark bottoms is one such common example. To curate a cleaner set of ground truth data, we create our latent SLAT with a ``de-lighted rendering'' by using ambient lighting to minimize such artifacts. We use this to process all stages of data. We select a uniform lighting strength, based on computing the similarities between \method predictions and the original image using RP-3DO (FO).

\paragraph{Pretraining.}
We start with pretraining the model on Iso-3DO-500K, a partition of Iso-3DO data with high aesthetics. Training on such data allows the model to predict plausible, high-quality texture that is characteristic of many 3D asset generation models. To ensure robustness of the texture model on real-world, complex images, we must further train on increasingly challenging data, via additional training stages described below.

For pretraining data in Iso-3DO, we render conditional images from the 3D mesh. We introduce a random lighting augmentation, where for each view, we apply a random lighting setting in the rendering engine. We hope the model can learn to remove the lighting effects when generating textures.

\paragraph{Mid-training.}
During mid-training, we train on RP-3DO (FO and OS-A). It is starting at this stage where we additionally provide full image conditioning to the model, as we believe contextual cues can help the model predict plausible textures, especially when the object is heavily occluded.
We show the effect of training on RP-3DO (FO), as well as the effect of further adding RP-3DO (OS-A) training data to this stage, in \Cref{fig:ablations_texture}.

We also introduce image data augmentation for RP-3DO (FO): \textbf{Mask} augmentation and \textbf{Blur} augmentation. The Mask augmentation randomly erode or dilate the input mask. This is designed to handle noise in mask at inference time (\eg segmentation predictions from a model). The Blur augmentation applies a downsample operation on the image followed by an upsample operation. This is especially important to handle cases for motion blur and small objects in an image. These augmentation is carefully studied in~\Cref{sec:texture_eval}.

\paragraph{Supervised fine-tuning (SFT).}
In the SFT stage, the model is trained on MITL-3DO texture annotations, which includes the ``aesthetic'' samples described in \Cref{sec:data_eng_img_obj_selection}. We show the effect of scaling the MITL-3DO texture annotations by 2x in \Cref{fig:ablations_texture}, which improves human preference rate by 14.2\%.

\paragraph{Preference optimization.}
Like the Geometry model, we run a final DPO stage to align the model with human preferences collected from the texture data engine. The effect of DPO on texture performance is shown in both \Cref{table:stage} and \Cref{fig:ablations_texture}. We follow the same training objective described in Equation~\eqref{eq:dpo}.

\subsection{Texture \& Refinement VAE}
\label{sec:vae_improvements}
We make improvements over the original SLAT VAE design in \citet{xiang2025structured}, where features are back-projected to all voxels, including those that are not visible (\ie, occluded) from the current image. This original design choice leads to reduced sharpness in the reconstructed images. To address this issue, we back-project features only to voxels that are visible from each image, utilizing the depth information from that specific view. We call this VAE variant Depth-VAE. During training, we normalize the Kullback–Leibler (KL) regularization term by the active voxel count to prevent large objects from dominating the training loss. We also fine-tune the decoder for downstream needs, such as reducing the number of decoded Gaussians for faster inference.  

\subsubsection{Depth-VAE: Depth-Aware Feature Aggregation}
To integrate depth information into patch-level features, we propose a depth-guided projection algorithm. Given a feature map $\mathbf{F} \in \mathbb{R}^{B \times C \times H \times W}$ and normalized 2D coordinates $\mathbf{U} \in [-1,1]^{B \times N \times 2}$, we sample features and aggregate them based on visibility, handling occlusions via a depth buffer.

\paragraph{Feature Sampling.}
For each coordinate $\mathbf{u}_i \in \mathbf{U}$, we extract the corresponding feature vector $\mathbf{f}_i$ from the DINO-V2 map $\mathbf{F}$ using differentiable bilinear interpolation (denoted as \texttt{GridSample}).

\paragraph{Occlusion Handling (Depth Filtering).}
To identify visible points, we construct a temporary depth buffer. We map the coordinates $\mathbf{U}$ to a discrete grid $(P_x, P_y)$ and retain the minimum predicted depth $\hat{d}$ at each location to form a surface depth map $\mathbf{D}_{\text{surf}}$:
\begin{equation}
    \mathbf{D}_{\text{surf}}(x, y) = \min_{i : (x_i, y_i) = (x, y)} \hat{d}_i.
\end{equation}
We then resample this map at the original coordinates $\mathbf{U}$ to obtain the reference surface depth $\mathbf{d}_{\text{ref}}$. Note that if ground-truth depth is available, it is used in place of $\mathbf{D}_{\text{surf}}$.

\paragraph{Visibility Masking.}
We compute a binary mask $\mathbf{M}$ to discard occluded points. A point is considered visible if its depth $\hat{d}_i$ is within a tolerance $\tau$ of the reference surface $\mathbf{d}_{\text{ref}, i}$:
\begin{equation}
    \mathbf{M}_i = \mathbb{I}\left[ \mathbf{d}_{\text{ref},i} > \hat{d}_i - \tau \right].
\end{equation}
We normalize this mask across the batch dimension (or views) to obtain weights $\tilde{\mathbf{M}}$.

\paragraph{Weighted Aggregation.}
The final depth-aware representation is the weighted sum of visible patch features:
\begin{equation}
    \mathbf{F}_{\text{depth}} = \sum_{b} \tilde{\mathbf{M}}_b \odot \mathbf{f}_b.
\end{equation}

\subsection{Training Hyperparameters}
\label{sec:training_details}
\label{sec:texture_hyperparameters}
\begin{table*}[t!]
\centering
\label{table:training_hyperparameters}
\resizebox{\textwidth}{!}{
\begin{tabular}{lllllllll}
\toprule
\textbf{Training stage} & \textbf{Datasets} & \textbf{Condition input} & \textbf{Learning rate} & \textbf{Modality weights} & \textbf{\# Meshes} & \textbf{\# Images} & \textbf{\# Samples} & \textbf{\# Tokens} \\
\midrule
Pre-training & Iso-3DO & object-centric crop & $10^{-4}$ & $S{=}1.0, R{=}0.1$ & 2.7M & 64.8M & 64.8M & 2.5T \\
\multirow{2}{*}[0mm]{Mid-training}
    & RP-3DO (FO)& full image & $10^{-4}$ & $S{=}1.0, R{=}0.1$ & 2.87M & 11.17M & 55.1M & 2.4T \\
    & RP-3DO (OS-R), ProcThor & full image, pointmap & $10^{-4}$ & $S{=}1.0, R{=}0.1, t{=}1.0, s{=}0.1$ & 2.38M & 1.20M & 5.95M & 0.3T \\
SFT & MITL-3DO, Art-3DO & full image, pointmap & $10^{-5}$ & $S{=}1.0, R{=}0.1, t{=}1.0, s{=}0.1$ & 0.6M & 0.5M & 0.6M & 0.9T \\
Alignment & MITL preference & full image, pointmap & $2.5\times10^{-6}$ & $S{=}1.0$ & 88K & 31K & 44K & - \\
\bottomrule
\end{tabular}
}
\caption{\textbf{Detailed training hyperparameters for SAM 3D training stages (Geometry Model)}. This table extends Table 1 from the main paper with additional hyperparameter details.}
\label{table:geometry_model_training_details}
\end{table*}
\begin{table*}[t!]
\centering
\label{table:texture_training_hyperparameters}
\resizebox{\textwidth}{!}{
\begin{tabular}{lllllllll}
\toprule
\textbf{Training stage} & \textbf{Datasets} & \textbf{Condition input} & \textbf{Learning rate} & \textbf{EMA} & \textbf{\# Meshes} & \textbf{\# Images} & \textbf{\# Samples} & \textbf{\# Tokens} \\
\midrule
Pre-training & Trellis500K & object-centric crop & $10^{-4}$ & 0.9999 & 350K & 9M & 10M & 1.1T \\
Mid-training & RP-3DO (FO,OS-A) & full image & $10^{-4}$ & 0.9999 & 800K & 2.4M & 2.4M & 1T \\
SFT & MITL & full image & $10^{-5}$ & 0.999 & $\sim$100K & 100K & 100K & 115B \\
Alignment & MITL preference & full image & $10^{-6}$ & 0.99 & 146K & 73K & 73K & - \\
\bottomrule
\end{tabular}
}
\caption{\textbf{Detailed training hyperparameters for SAM 3D training stages (Texture \& Refinement Model).} This table extends Table 1 from the main paper with additional hyperparameter details.}
\label{table:texture_model_training_details}
\end{table*}

We summarize training parameters in details in~\Cref{table:geometry_model_training_details} for Geometry model and~\Cref{table:texture_model_training_details} for Texture \& Refinement model.

We use a batch size of 6 per GPU for all training stages of the Geometry model, and epochs iterate over meshes. Pretraining is conducted on 512 A100 GPUs for 200 epochs. Mid-training on FO utilizes 320 A100 GPUs for 50 epochs, followed by further FO mid-training on 128 A100 GPUs for an additional 50 epochs, followed by OS-R midtraining on 256 A100s for 12 epochs. SFT is performed on 128 H200 GPUs for 100 epochs, training on data from our data engine as it becomes available. As this data leads to model improvements (and thus also improving the quality of data produced by the data engine), we raise our quality threshold $\alpha_k$ for keeping samples in our SFT training set; the final run uses an quality cutoff $\alpha_K$ that keeps $500$K samples. DPO is performed on 128 A100 GPUs for 1 epoch.

For the Texture \& Refinement model, we perform pretraining on 256 A100s for 245 epochs with a batch size of 4, followed by mid-training on 256 A100s for 80 epochs with a batch size of 4. For SFT, we use 192 A100s for 89 epochs and batch size of 4. Finally, DPO is conducted on 128 A100s for 2 epochs with a batch size of 3.

\section{Evaluation}
\label{section:eval}

Current evaluation benchmarks for visually grounded 3D object reconstruction fall short of capturing the complexity of the real world. Many rely on synthetic datasets~\citep{deitke2023objaverse,chang2015shapenet} where single objects are rendered in isolation, centered against a white background. This introduces a large visual gap with real-world evaluation conditions and the rich variation of real-world imagery. Efforts to move to real data mostly focus on indoor environments~\citep{khanna2024habitat,sun2018pix3d,pan2023aria}, but these benchmarks heavily skew toward furniture categories such as chairs and tables, limiting the diversity of objects that models must handle in practice. As a result, evaluations on these datasets do not reflect the challenges of natural environments, where objects are occluded, scenes are cluttered, scales vary, lighting conditions complicate appearance, and domains span far beyond indoor or synthetic scenes.

\subsection{SA-3DAO: A New Benchmark for Real-World 3D Object Reconstruction}
\label{sec:sa3dao}
We introduce SAM 3D Artist Objects (SA-3DAO), a new benchmark designed to capture the diversity and complexity of real-world 3D perception. SA-3DAO consists of 1,000 untextured 3D objects, created from and carefully aligned to selected natural images capturing scenes spanning both indoor and outdoor environments, including parks, ski resorts, flea markets, parades and more. The benchmark covers a wide spectrum of object types: objects range from large, structured entities such as ski lifts and escalators, to everyday items like clothing, to rare and culturally specific objects such as tribal face masks. Crucially, the 3D ground truth are of high-fidelity created by professional 3D artists, who are tasked with producing accurate 3D shapes for the objects depicted in the input images. 
This combination of visual diversity, real-world context, and professionally crafted 3D ground truth makes SA-3DAO a comprehensive testbed for evaluating 3D object reconstruction models.

\begin{wrapfigure}{r}{0.5\columnwidth}
  \centering
  \vspace{-4mm}
  \includegraphics[width=0.5\columnwidth]{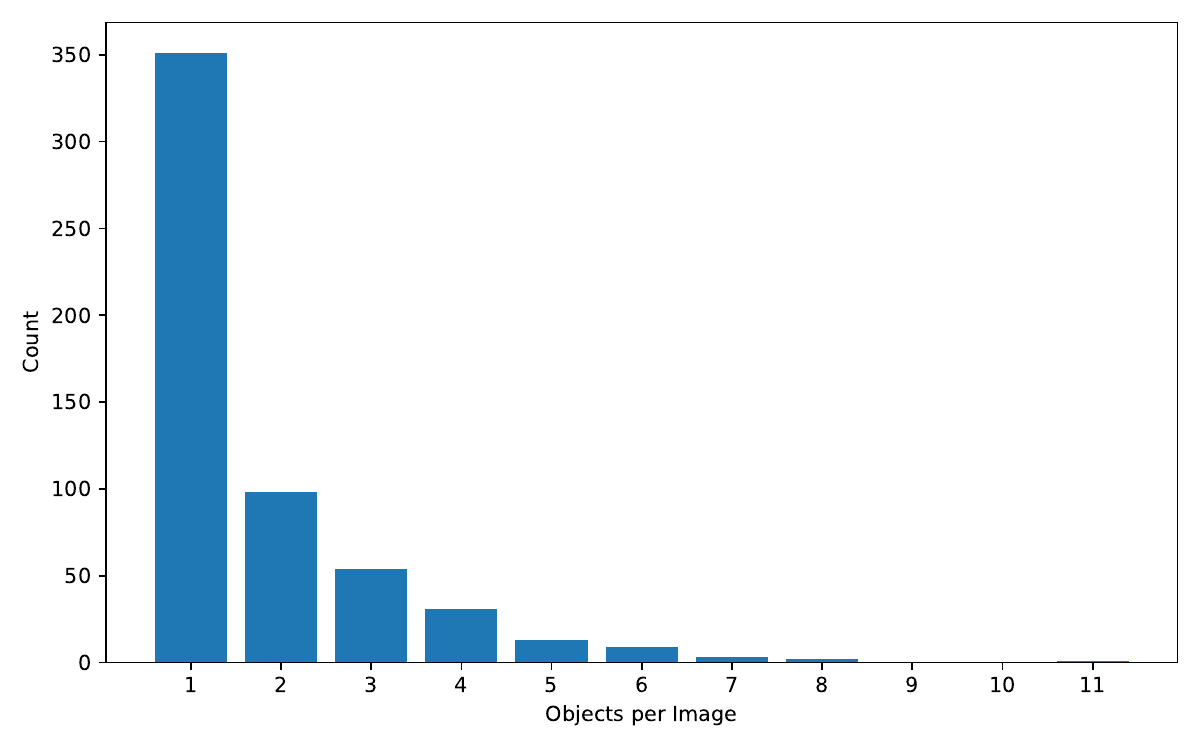}
  \caption{\textbf{Distribution of number of objects per image in SA-3DAO.} The number of objects follows a roughly power-law distribution.}
  \label{fig:3dao_obj_freq}
  % \vspace{-4mm}
\end{wrapfigure}
\paragraph{Collection details.} We task professional 3D artists with recovering the shape of a target object from a single image, mirroring our model's goal of reversing the photographic transform (as described in \Cref{sec:problem_formulation}).
In other words, the 3D artists must create a whole 3D mesh that precisely aligns with the object's visible pixels in the image.
Even under ideal settings, this requires contending with only partial information as the back side of the object is typically unseen, but many of the objects in SA-3DAO additionally have natural occlusions, or are small in size within the image; disentangling depth versus scale is often also challenging for a single image.
To fill these information gaps, artists rely on recognition and context, using common-sense priors, physical plausibility, and assumptions of symmetry (when appropriate) to complete the meshes.
The requirements imposed by this task are atypical to the normal 3D asset creation process that artists are more accustomed to, and efficient annotation requires learning a different mode of operation.
After acclimation to the task, completion time per object mesh can vary considerably, ranging from up to 5 minutes for obvious objects with simple geometries to over 5 hours for more challenging cases; the median mesh in our dataset has 4751 vertices.
Many of the images provided multiple objects with meshes from the 3D artists; we show the frequency distribution in \Cref{fig:3dao_obj_freq}.

\subsection{Human Preference Set}
We further expand our evaluation suite to support more rigorous and domain-targeted assessments. While SA-3DAO provides a general and standardized way to measure progress, we want to also capture the challenges of settings where 3D perception is most critical, such as robotic manipulation and egocentric vision. To address this, we design a human preference set composed of images drawn from these domains of interest. This set enables evaluation through direct human judgment, providing insights that go beyond numerical metrics and capturing aspects of 3D perception that are important in embodied and real-world applications. 

\paragraph{Domains.} 
We design four human preference test datasets to comprehensively evaluate model capabilities across different scenarios. 
\begin{itemize}
    \item \textbf{SA-1B}~\citep{kirillov2023segment}: We uniformly sample 1,000 image and object mask pairs, covering a diverse range of object categories. This set is intended to assess the model’s generalization ability across varied object distributions, with particular emphasis on long-tail categories.
    \item \textbf{MetaCLIP}~\citep{xu2024metaclip}: We select 1,000 samples where the object masks are of median or heavily occluded. This set evaluates the model's performance in reconstructing occluded objects, a common challenge in cluttered scenes. 
    \item \textbf{LVIS}~\citep{gupta2019lvis}: We densely sample 1,000 images containing between 10 and 30 objects per scene, and is designed to evaluate the model’s transferability to out-of-domain data and to demonstrate its ability to capture physical properties within dense scene layouts. 
    \item \textbf{Aria Digital Twin}~\citep{pan2023aria}: We sample a smaller set of ~40 video frames, with around 30 objects per scene. This dataset is intended to compare against baselines on scenes with highly accurate pointmaps, similar to those on which the baselines were trained. 
\end{itemize}

\paragraph{Setup.} Human preference evaluations are conducted through a structured sequence of pairwise comparisons. For each image and a masked object, annotators are first presented with two reconstructions (model ``A'' vs. ``B'') and asked to select the one that most accurately matches the object in the image. The chosen reconstruction is then compared against the output of a third model (model ``C''), and this process continues iteratively until all candidate models have been compared. Through this series of binary decisions, the most accurate reconstruction is identified as the preference for that particular image. To ensure fairness and avoid bias, the order of comparisons is randomized and the identity of models are anonymized.

\subsection{Evaluation Metrics}
\subsubsection{Shape Metrics Definitions}
For shape evaluation on SA-3DAO, we first normalize the artist-created ground-truth mesh and the generated mesh independently into the range $[-1,1]$. We then apply ICP alignment for each mesh pair before computing metrics. We uniformly sample $1$M points from both meshes and report four complementary metrics that capture different aspects of geometric fidelity:

\begin{itemize}
    \item \textbf{F-score @ 0.01:}
    Measures correspondence accuracy between the reconstructed and ground-truth points under a $0.01$ threshold. We compute precision and recall between the two point clouds and report their harmonic mean. F1 evaluates how many points lie close to the ground truth and how completely the reconstruction covers the target shape.

    \item \textbf{Voxel-IoU:}
    Provides a coarse volumetric agreement score and is sensitive to gross errors in volume, silhouette, and topology.
    We voxelize both point clouds to $64^3$ resolution and compute intersection-over-union over occupied voxels.

    \item \textbf{Chamfer Distance (CD):} 
    Measures bidirectional nearest-neighbor distance between reconstructed and ground-truth point sets, highlighting fine-grained geometric deviation and penalizing missing or distorted regions.

    \item \textbf{Earth Mover’s Distance (EMD):} 
    Quantifies the minimal cost required to transport one point distribution to match the other. EMD is more stringent than CD, capturing global structural differences and enforcing bijective correspondence between distributions.
\end{itemize}
Together, these metrics provide a comprehensive view of reconstruction fidelity, from local accuracy to global shape consistency.

Moreover, to evaluate shape quality on the ISO3D dataset~\citep{ebert20253d}—which consists of $101$ in-the-wild synthetic images without 3D ground truth—we measure perceptual similarity between the generated shape and the input image using ULIP~\citep{xue2023ulip} and Uni3D~\citep{zhou2023uni3d}. For each generated mesh, we uniformly sample $8,192$ surface points to form a point cloud representation, and compute cross-modal similarity between the point cloud features and image features.

\subsubsection{Layout Metrics Definitions}
To evaluate single-object pose and compare with existing methods, we employ standard 6D pose estimation metrics, and then define ICP rotation error below.

\begin{itemize}
    \item \textbf{3D IoU:}
    Measures the overlap of 3D axis-aligned bounding boxes between predicted and ground-truth  bounding boxes, using the intersection-over-union. Values range from 0 (no overlap) to 1 (perfect overlap).
    
    \item \textbf{ICP-Rot:} \emph{ICP Rotation Error} is
    the residual rotation error (in degrees) after ICP alignment. Given predicted rotation ${R}_{\text{pred}}$ and ground-truth rotation ${R}_{\text{gt}}$, the meshes are first posed, then ICP finds optimal alignment ${R}_{\text{ICP}}$, and \textbf{ICP-Rot} is the angle of this rotation in degrees.
     
    \item \textbf{ADD-S (Average Distance with Symmetry):}
    ADD-S~\citep{xiang2018posecnn} is the symmetrized average of the minimum point-to-point distances between predicted and ground-truth posed objects, normalized by object diameter:
    \begin{align}
    \text{ADD}(\mathcal{A}, \mathcal{B}) &= \frac{1}{|\mathcal{A}|} \sum_{\mathbf{x} \in \mathcal{A}} \min_{\mathbf{y} \in \mathcal{B}} \|\mathbf{x} - \mathbf{y}\|_2 \\
    \text{ADD-S} &= \frac{\text{ADD}(\mathcal{M}, \mathcal{M}_{\text{gt}}) + \text{ADD}(\mathcal{M}_{\text{gt}}, \mathcal{M})}{2d}
    \end{align}
    where $\mathcal{M}$ and $\mathcal{M}_{\text{gt}}$ are the predicted and ground-truth point clouds for the posed shape, and $d = \max_{\mathbf{x}, \mathbf{y} \in \mathcal{M}_{\text{gt}}} \|\mathbf{x} - \mathbf{y}\|_2$ is the diameter of the ground-truth point cloud. The symmetrized formulation averages distances in both directions: from predicted to ground-truth and from ground-truth to predicted. Lower values indicate better pose accuracy.
    
    The original ADD-S metric definition~\citep{xiang2018posecnn} was designed for 6DoF pose estimation using a ground truth CAD model. In this case, when the predicted and ground truth shape are the same, the asymmetric and symmetric versions of ADD-S coincide. In \method we jointly estimate shape and pose, so generalize the metric to the symmetric version.
    
    \item \textbf{ADD-S @ 0.1:}
    A binary value per-sample indicating whether the ADD-S distance is less than 10\% of the object's diameter. 
\end{itemize}

\section{Additional Ablations}
\subsection{Intermediate Training Stage Knockout}
\label{sec:mitl_ablation}

While~\Cref{table:stage} in the main paper shows the cumulative effect of adding different stages during training, \Cref{table:ablate_stage} shows the impact of real-world data as intermediate stages. Knocking out any of these stages results in a notable drop in shape performance.

\begin{table}[ht!]
    \centering
        \begin{NiceTabular}{llcccc}
        \CodeBefore
        \Body
        \toprule
        &   & \multicolumn{4}{c}{SA-3DAO} \\
        \cmidrule(lr){3-6}
        Model  &  Training Setup   & F1 @ 0.01 ($\uparrow$) & vIoU ($\uparrow$)  & Chamfer ($\downarrow$) & EMD ($\downarrow$) \\
        \midrule
        \method & Full & \textbf{0.2344} & \textbf{0.2311} & \textbf{0.0400} & \textbf{0.1211} \\
                & w/o training on MITL-3DO & 0.2211 & 0.2220 & 0.0486 & 0.1338 \\
                & w/o training on Art-3DO & 0.2027 & 0.2025 & 0.0578 & 0.1510 \\
                & w/o DPO on MITL-3DO & 0.2156 & 0.2156 & 0.0498 & 0.1367 \\
        \bottomrule
        \end{NiceTabular}
    \caption{\textbf{Training stage knockout.} The impact of training on MITL and 3D artist-generated data.}
    \label{table:ablate_stage}
\end{table}

\subsection{Texture Evaluations}
\label{sec:texture_eval}
\begin{figure}[ht!]
    \centering
    \begin{minipage}[c]{0.48\textwidth}
        \centering
        \resizebox{\linewidth}{!}{
            \begin{NiceTabular}{lcccc}
            \CodeBefore
            \Body
            \toprule
            & \multicolumn{4}{c}{\method WR over baselines, \method shape} \\
            \cmidrule(lr){2-5}
            Model & ISO3D & Preference Set & SA-3DAO & LVIS \\
            \midrule 
            Trellis & 81.1 & 87.0 & 86.2 & 89.1 \\
            Hunyuan3D-2.1 & 63.8 & 87.0 & 86.2 & 89.1 \\
            Hunyuan3D-2.0 & 70.1  & 77.5 & 77.4 & 85.7 \\
            Unitex & 83.3 & 84.7 & 84.5 & 88.3 \\
            \bottomrule
            \end{NiceTabular}
        }
        \captionof{table}{\textbf{3D texture.} Preference results comparing \method to competing image-to-3D methods on ISO3D~\citep{ebert20253d}, Preference Set, and SA-3DAO. We compare to the recent Trellis~\citep{xiang2025structured}, Hunyuan3D-2.1~\citep{hunyuan3d2025hunyuan3d}, and Unitex~\citep{liang2025unitex}, with the same shape of \method. The human preference rates represent preference for \method over each baseline approach.}
        \label{table:comparison_texture}
    \end{minipage}
    \hfill 
    \begin{minipage}[c]{0.48\textwidth}
        \centering
        \includegraphics[width=\linewidth]{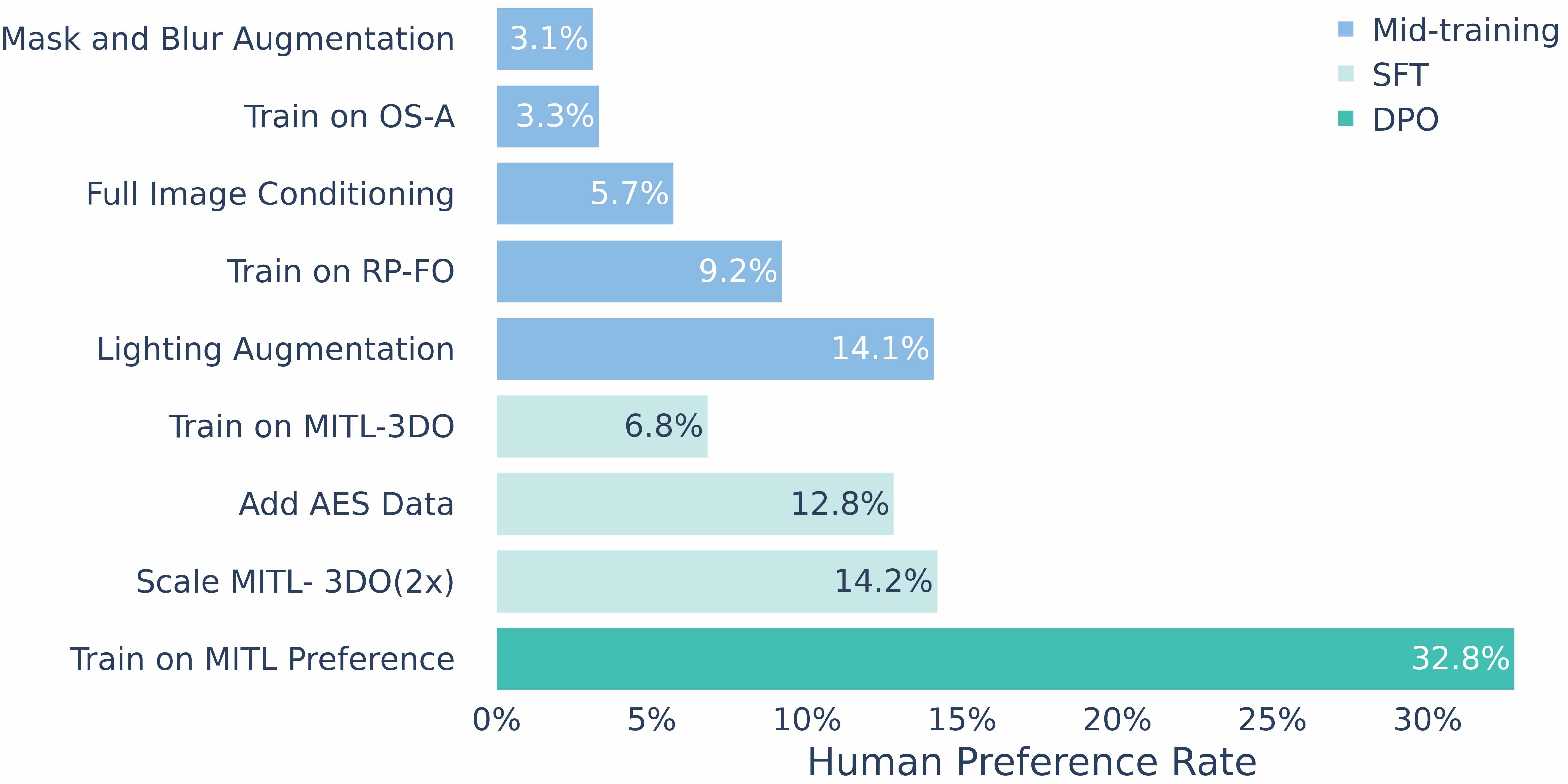}
        \caption{\textbf{Ablations for the Texture \& Refinement model, grouped by training stage.} Percentages denote human preference rate for each ablation, over the model \textit{without} the ablation.}
        \label{fig:ablations_texture}
    \end{minipage}
\end{figure}

\paragraph{Comparison to SOTA.} We compare \method with existing methods on a holistic level (combined geometry and texture prediction) in \Cref{table:comparison_texture}.
We compare against existing image-to-3D methods that predict Gaussians or textured meshes, including Trellis~\citep{xiang2025structured} and Hunyuan3D-2.1~\citep{hunyuan3d2025hunyuan3d}.
We also conduct a texture-only comparison by providing \method geometry as input to the texture modules of the aforementioned baselines, with the addition of Unitex~\citep{liang2025unitex}, a model that performs texture prediction given paired image and shape input.

We report human preference for \method over each baseline method on multiple datasets: ISO3D~\citep{ebert20253d}, Preference Set, SA-3DAO, and LVIS~\citep{gupta2019lvis}. Results indicate that \method outperforms existing methods on the holistic image-to-3D task as well as texture estimation -- this is due to the fact that the evaluated datasets often contain occlusion and clutter, which is a setting prior works struggle on.

\paragraph{Ablations for Texture Training.} We conduct comprehensive studies on design choices for Texture \& Refinement model (\Cref{fig:ablations_texture}) using annotator preferences on the Pref Set. We benchmark each component to the alternative model without the change. We remark a few themes here:

\begin{itemize}
    \item Augmentation is very important, with lighting augmentation to be the most critical here. This is expected, given the Mask and Blur augmentations primarily focus on specific challenging cases (poor mask quality and low resolution inputs), so their effects get diluted in a holistic evaluation. 
    \item RP-3DO data are critical and helps the model adapt to real world. 
    \item Post-training data are critical, with significant gains coming from it. It demonstrates the effectiveness of our Data Engine, and DPO further amplifies the gains. In addition, sourcing specific type of data (AES) and scaling the data both show significant improvements. 
\end{itemize}

\subsection{Layout Test-Time Optimization}
\label{sec:post-optim}
Render-and-compare is a longstanding popular approach for pose estimation~\citep{labbe2022megapose, wen2024foundationpose}: iteratively render the object's shape according to the most recently predicted pose, directly compare with the input pixels, and then adjust the pose prediction accordingly.
This heuristic-based search can lead to fairly accurate results, even with weak initial pose samples (``proposals'' from a base model).
By contrast, \method operates in a feedforward manner, directly diffusing the object pose (rotation, translation, and scale) conditioned on the image features and, optionally, the scene pointmap.
Notably, we do not include any sort of pixel-based loss objective in this process.
However, \method's pose estimation can very naturally be used as a proposal for render-and-compare optimization.
We include experiments showing the impact of this post-optimization process in \Cref{table:comparison_post-optim}, demonstrating that we can achieve further gains in pose metrics on ADT~\citep{pan2023aria}. 

  \begin{table*}[h]
      \centering
      \resizebox{0.7\textwidth}{!}{
          \begin{NiceTabular}{lccccc}
          \CodeBefore
          \Body
          \toprule
          Model   & 3D IoU ($\uparrow$) & ICP-Rot. ($\downarrow$) & ADD-S ($\downarrow$) & ADD-S @ 0.1 ($\uparrow$) & 2D IoU ($\uparrow$) \\
          \midrule
          \method & 0.4837 & 14.4702 & 0.08265 & 0.7545 & 0.5143 \\
          \method~(post-optim) & \textbf{0.5258} & \textbf{14.1460} & \textbf{0.07932} & \textbf{0.7617} & \textbf{0.6487} \\
          \bottomrule
          \end{NiceTabular}
      } 
      \caption{\textbf{Test-time optimization for layout.} Quantitative comparison of \method layout test-time optimization on Aria Digital Twin~\citep{pan2023aria}.}
      \label{table:comparison_post-optim}
  \end{table*}

Specifically, we further optimize the layout proposals from \method by applying the layout to the generated objects, rendering and comparing for both masks and pixels, and backpropagating the gradients to refine the layout. We finally apply an automatic proposal-checking step, where the optimized layout is accepted only if its mask IoU exceeds that of the initial layout. As shown in \Cref{table:comparison_post-optim}, for the $554$ accepted optimized samples out of the $1027$ ADT instances, both the 3D layout metrics and the 2D mask IoU metric improve substantially, demonstrating the effectiveness of layout test-time optimization.

\subsection{Rotation Representation}
\label{sec:rot_representation_ablation}
We compare different rotation representations while keeping all other architecture and training settings identical. Each model is trained on the pretraining dataset and evaluated on a held-out Objaverse test split of 216 samples. As shown in \Cref{tab:rotation_ablation}, switching from a quaternion representation to the 6D continuous rotation parameterization~\citep{zhou2019continuity} yields a notable reduction in oriented rotation error, confirming that the 6D formulation provides a smoother optimization landscape more suitable for generative modeling. Further applying normalization to the 6D rotation vectors using the statistics over the training datasets leads to an additional improvement when training the flow matching models.

\begin{table}[h]
\centering
\resizebox{0.45\textwidth}{!}{
\begin{NiceTabular}{lcc}
\CodeBefore
\Body
\toprule
\textbf{Representation} & \textbf{Chamfer} ($\downarrow$) & \textbf{ICP-Rot.} ($\downarrow$) \\
\midrule
Quaternion              & 0.0061 & 17.9585 \\
6D Rotation             & 0.0074 & 15.5399 \\
Normalized 6D Rotation  & \textbf{0.0049} & \textbf{14.5946} \\
\bottomrule
\end{NiceTabular}
}
\caption{\textbf{Rotation representation.} Ablation on the representation used during pretraining. We report Chamfer distances and ICP rotation error.}
\label{tab:rotation_ablation}
\end{table}

\subsection{Pointmap Minimally Affects Shape}
\label{sec:rgb_only_model_ablation}
\method can condition on a 2.5D pointmap derived from sensor measurements (see~\Cref{sec:architecture}) or from an off-the-shelf monocular depth estimation derived from the image itself, as is primarily used throughout this work. The latter is notable, as it also means that the current model can continue to benefit from future improvements of depth estimation methods.
We observe that the pointmap minimally affects the shape performance: in a head-to-head preference test for shape on LVIS, the version of \method conditioned on pointmaps and the version without pointmaps are each selected 48\% of the time.

\subsection{Texture \& Refinement \emph{Depth-VAE} Comparison}
\label{sec:depthVAE_ablation}

\Cref{tab:depth-vae} shows the result of the improvements made to the VAE used in the Texture \& Refinement model; see ~\Cref{sec:vae_improvements}. We found that the depth feature significantly improves the perceptual quality of reconstruction, while scaling the training data further improves the reconstruction performance. We also notice that the enhancement primarily arises from the difficult scenarios for the non-depth VAE (when it has a bad performance).

\begin{table}[h]
\centering
\resizebox{0.45\textwidth}{!}{
\begin{NiceTabular}{lccc}
\CodeBefore
\Body
\toprule
\textbf{Method} & \textbf{PSNR} ($\uparrow$) & \textbf{SSIM} ($\uparrow$) & \textbf{LPIPS} ($\downarrow$)\\
\midrule
Non-Depth VAE              & 30.65 & 0.9470 & 0.04776\\
Depth-VAE             & 30.87 & 0.9500 & 0.04579\\
Depth-VAE + scaling  & \textbf{31.60} & \textbf{0.9547} & \textbf{0.04093} \\
\bottomrule
\end{NiceTabular}
}
\caption{\textbf{Depth-VAE ablations.} Effectiveness the depth-feature modification to the SLAT VAEs used in the Texture \& refinement model. Results are evaluated on the entire GSO dataset.}
\label{tab:depth-vae}
\end{table}

\subsection{Data Engine: Increasing Best-of-$N$ Search with Reward Models}
\label{sec:best_of_n_ablation}
\begin{table}[h]
\centering
\resizebox{0.7\columnwidth}{!}{
\begin{tabular}{l|cc|cc|cc}
\toprule
& \multicolumn{2}{c|}{Tail Holdout} & \multicolumn{2}{c|}{Epic Kitchens} & \multicolumn{2}{c}{SA-3DAO} \\
& Chamfer $\downarrow$ & F1 $\uparrow$ & Chamfer $\downarrow$ & F1 $\uparrow$ & Chamfer $\downarrow$ & F1 $\uparrow$ \\
\midrule
SFT with $N=2$ & 0.0059 & 0.39 & 0.0094 & 0.30 & 0.0083 & 0.26 \\
SFT with $N=50$ recovery & \textbf{0.0053} & \textbf{0.41} & \textbf{0.0090} & \textbf{0.32} & \textbf{0.0081} & \textbf{0.26} \\
\bottomrule
\end{tabular}
}
\caption{\textbf{Including reward-model-recovered data during SFT}, from the pipeline in \Cref{fig:vlm_search}, improves model performance on challenging inputs, as seen in both Chamfer Distance and F1 score on the tail holdout set, Epic Kitchens~\citep{damen2020epic}, and SA-3DAO.}
\label{table:vlm_search}
\end{table}
Finetuning on data recovered using the reward-model-best-of-$N$ pipeline improves model performance on various challenging inputs, such as the artist evaluation set, tail holdout set, as well as Epic Kitchens~\citep{damen2020epic}, as shown in \Cref{table:vlm_search}. This demonstrates that further amplifying the expert policy in~\Cref{alg:data_engine} by increasing $N$ in the best-of-$N$ search can improve the robustness of the model in challenging categories, and suggests that improved test-time search can increase the alignment convergence speed of the data engine.

We found that both vision-language models (VLMs), and also the implicit reward models from our DPO stage~\citep{lambert2024rewardbenchevaluatingrewardmodels} performed similarly in our case. In our testing, the VLM-as-reward model had $68.9\%$ binary choice agreement with humans rater preferences, DPO agreed $\sim65\%$, and two human annotators agreed $<75\%$\footnote{While some of this may be from human error, we attribute most of this disagreement to random chance when the MITL suite returns multiple similar meshes, or when there is no clear winner due to meshes being better or worse in different parts.} of the time. Around 80\% of the recovery data came from the DPO-as-reward model.

\subsection{Model Distillation Results}
\begin{figure}[t!]
\centering
\begin{subfigure}[b]{0.94\textwidth}
\centering 
\includegraphics[width=\textwidth]{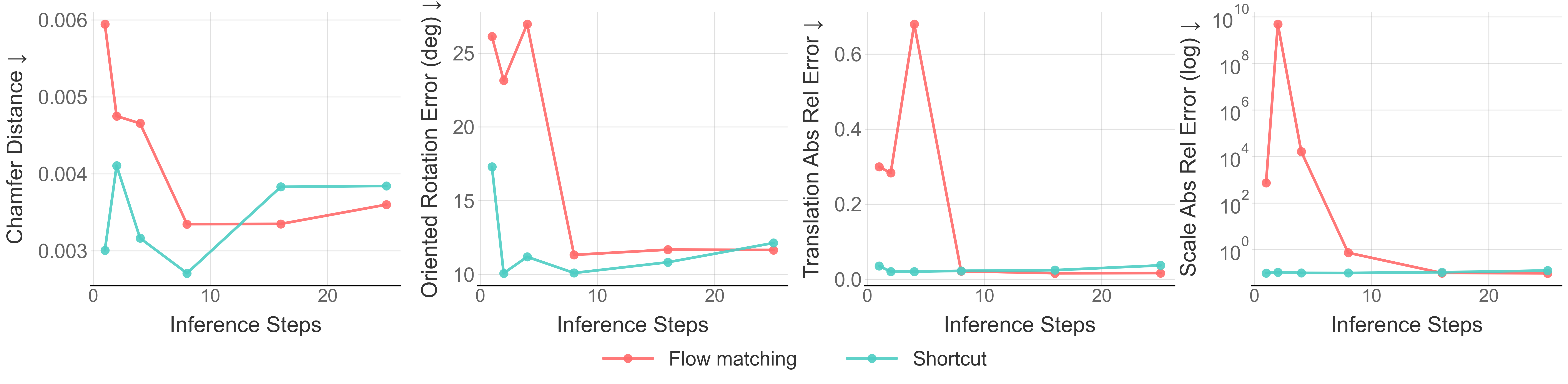}
\label{fig:preference:subfig1}
\end{subfigure}
\vspace{-0.5cm}
\caption{\textbf{Model distillation.} Geometry model shortcut versus flow matching. Flow matching distillation enables the model to perform significantly better during the early iterations, nearly on par with 25 steps performance.} 
\label{fig:distillation}
\end{figure}
\begin{figure*}[t]
\centering 
\includegraphics[width=\textwidth]{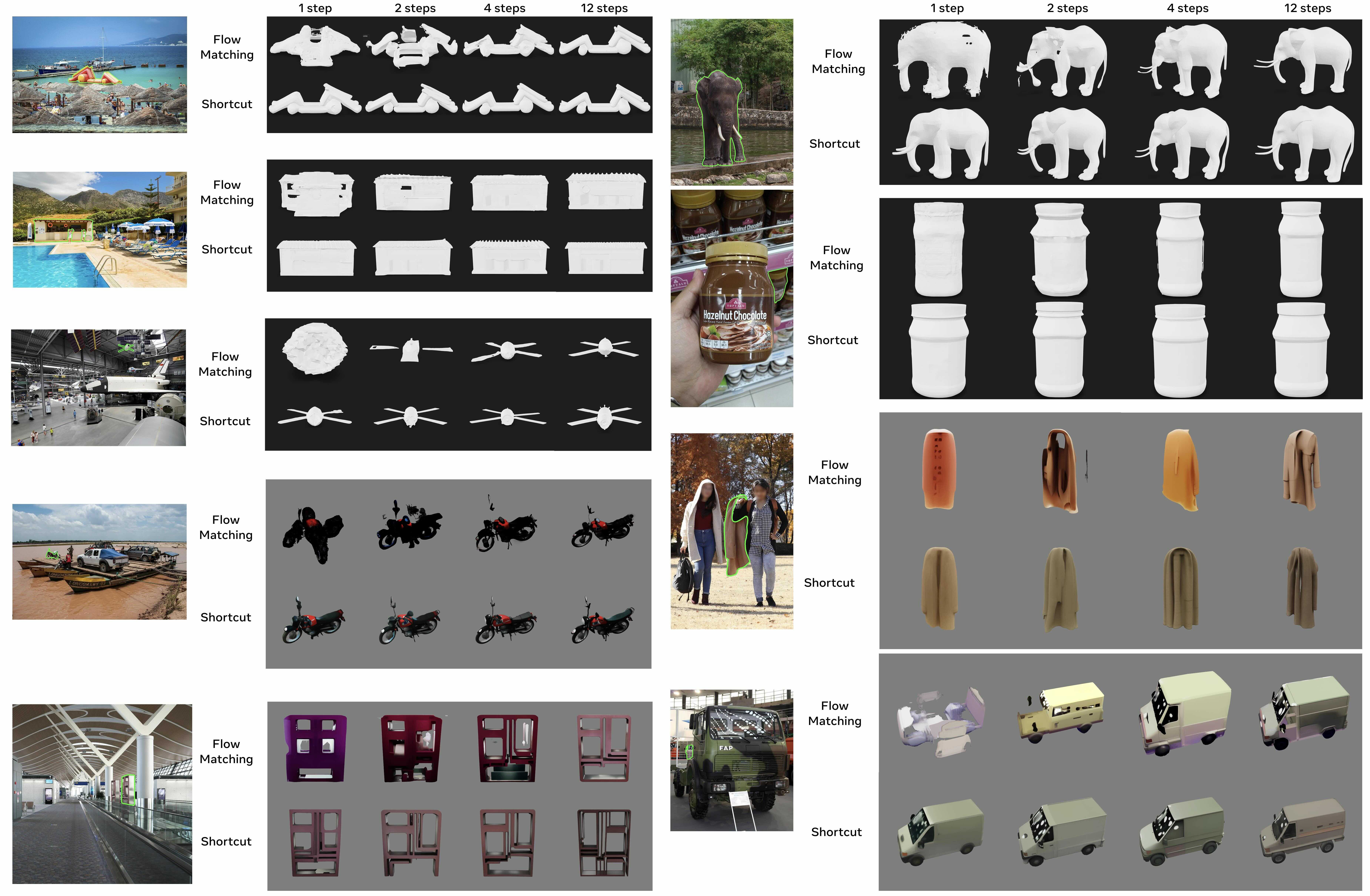}
\caption{\textbf{Qualitative examples after distillation.} Visualization using difference number of steps in flow matching mode and shortcut mode. The black background displays the mesh rendering without texture, while the grey background shows the rendering of the Gaussian splattings.}
\label{fig:distillation_steps_vis}
\end{figure*}
\Cref{fig:distillation} illustrates the performance of \method with and without distillation, plotted against the number of flow matching steps for the Geometry model. Specifically, using 1-step and 4-step methods yields a 38$\times$ and 10$\times$ inference speed improvement, respectively, compared to 25 steps with flow matching (w/ CFG). In flow matching mode, CFG is applied during the first half of the steps, resulting in a total NFE that is 1.5 times the number of steps. In contrast, shortcut mode achieves an NFE equal to the number of steps, as CFG is distilled directly into the model. For the Texture \& Refinement model, we opted not to apply distillation for the final release, as the final model already performs well with fewer steps out of the box. This is because the overall geometry is primarily determined by the Geometry model's voxel output, and increasing the number of steps does not significantly alter the geometry. However, as illustrated in~\Cref{fig:distillation_steps_vis}, where we present a visualization of the model outputs using different numbers of inference steps with shortcut mode enabled for both the Geometry model and the Texture \& Refinement model, shortcut model distillation improves texture quality when using fewer steps.

\subsection{Input Mask Robustness}
\begin{wrapfigure}{r}{0.5\columnwidth}
  \centering
  \vspace{-3mm}
  \includegraphics[width=\linewidth]{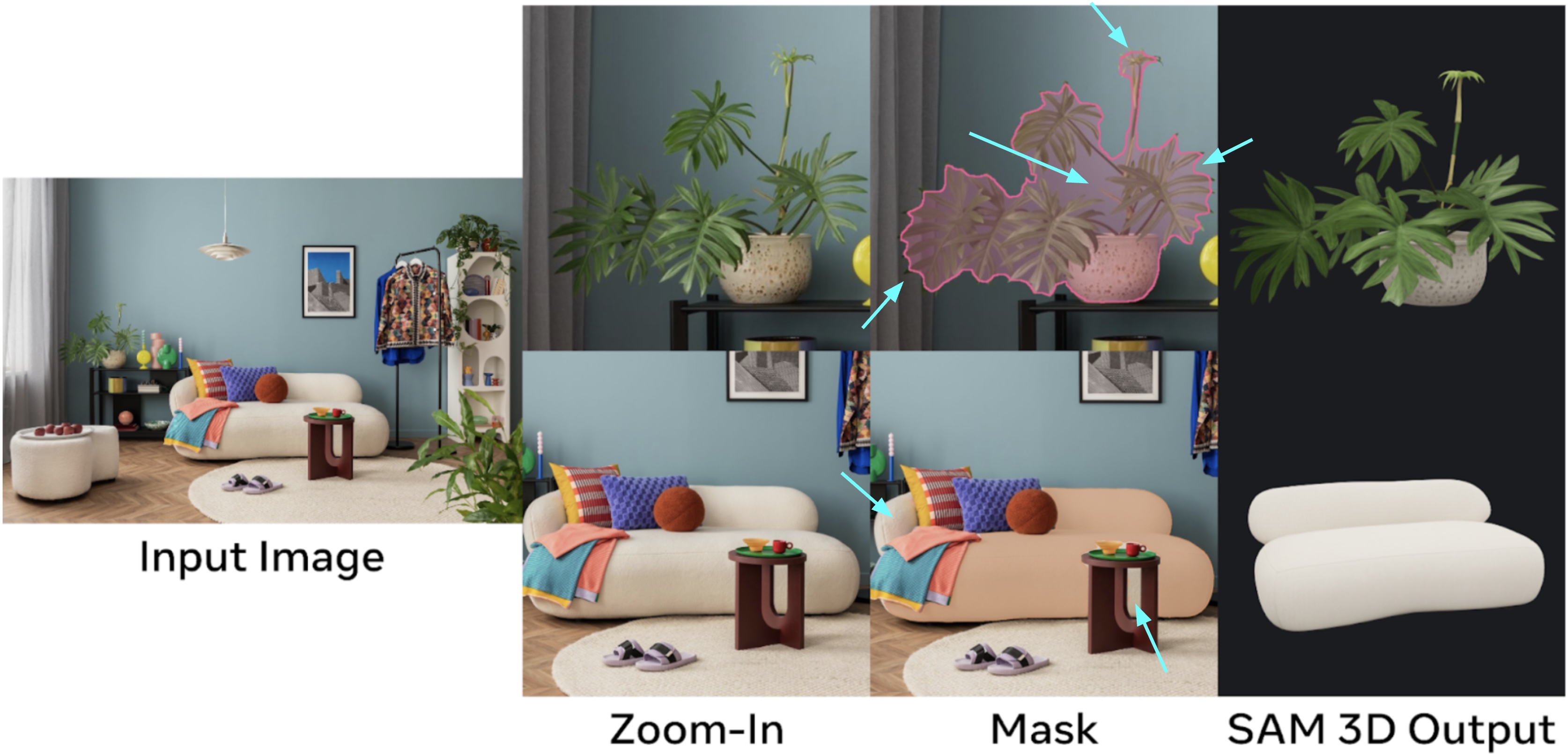}
  \vspace{-1mm}
    \caption{\textbf{SAM 3D robustness to input mask precision.} \textbf{Top}: Over-segmentation; \textbf{Bottom}: Under-segmentation.}
   \label{fig:over-under-seg}
\end{wrapfigure}

SAM 3D assumes a segmentation mask as input to designate to the model which object within a scene should be reconstructed, as well as isolate the object's pixels for visual feature extraction.
Accurate segmentation by hand can be time-consuming, but the ready availability of segmentation models like the SAM series~\citep{kirillov2023segment, ravi2024sam, carion2025sam} offer a fast, promptable solution.
On the other hand, automatic methods are not perfect, and as a preceding step to SAM 3D, there is risk of compounding errors. 
Yet, we observe empirically that SAM 3D has fair robustness to errors in mask precision (see~\Cref{fig:over-under-seg}), demonstrating the ability to recognize the intended target object and reconstruct it.
We also note that as an input to the model, SAM 3D is agnostic to the source of mask, and thus is able to take advantage of ground truth masks or advancements in segmentation by future models.

\section{Limitations}
There are limits to our model's resolution based on the architectural hyperparameters we used. 
The geometry model uses a coarse shape resolution of $O \in \mathbb{R}^{64^3}$; we trained multiple 3D Gaussian splat decoders, with up to $32$ splats per occupied voxel. 
This is sufficient for many types of objects, but for more complex shapes, thin structures, or where human perception is especially attuned, these limits to resolution can lead to noticeable distortions or loss of details.
For example, as part of a whole human body, the number of voxels/splats our chosen resolution is able to devote to hands or faces is inherently limited by the overall body's scale, and due to human visual system's acuity to such features, this can lead to perceptible artifacts to these body parts.
By contrast, when focused on just a single hand or the head, the higher relative resolution available means \method can reconstruct these significantly better.
For these kinds of objects and others, a natural next step for \method would be to increase the output resolution via architectural changes, a superresolution model, parts-based generation, or switching to an implicit 3D representation.

Object layouts are another area where improvements can be made. 
\method predicts objects one at a time, and isn't trained to reason about physical interactions, such as contact, physical stability, interpenetration, or co-alignment (\eg on the same ground plane). 
Multi-object prediction combined with appropriate losses would allow joint reasoning about multiple objects in a scene.
Similarly, while SAM 3D's single-object inference is embarrassingly parallel across multiple GPUs, scene-based inference may also present opportunities to more efficiently accelerate the reconstruction of entire scenes. 
Finally, \method's texture predictions are made without knowledge of the predicted object's pose; as a result, for objects with rotational symmetry, it can occasionally predict textures that effectively rotate the object to the incorrect orientation.

\begin{figure*}[t]
    \centering 
    \includegraphics[width=0.87\linewidth]{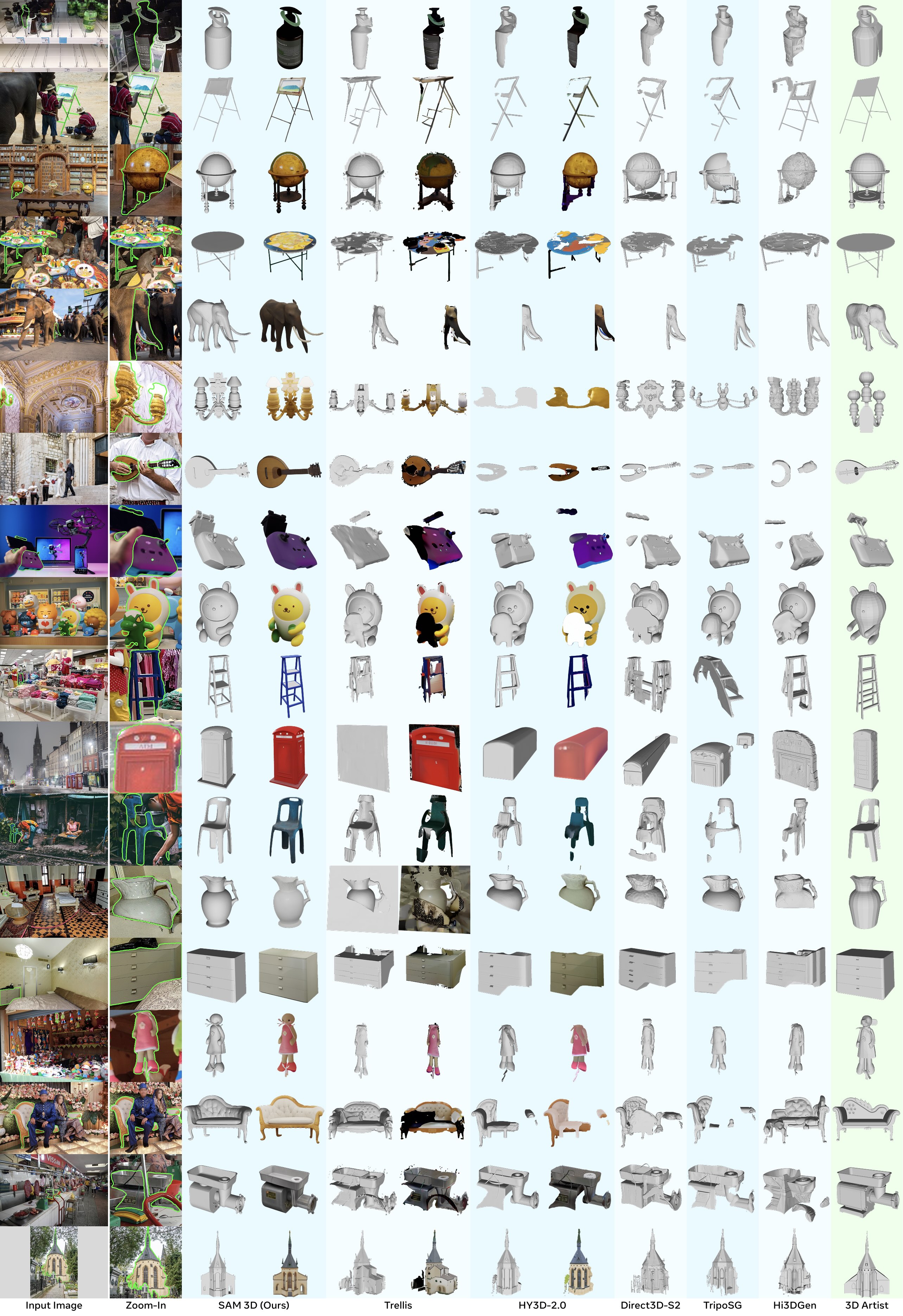}
    \caption{\textbf{Additional qualitative shape and texture results of our model on the SA-3DAO eval set.} For models that include texture, we show the untextured mesh (left) and textured mesh (right) separately.}
    \label{fig:comparison_shape_2}
\end{figure*}
\begin{figure*}[t]
    \centering 
    \includegraphics[width=0.85\linewidth]{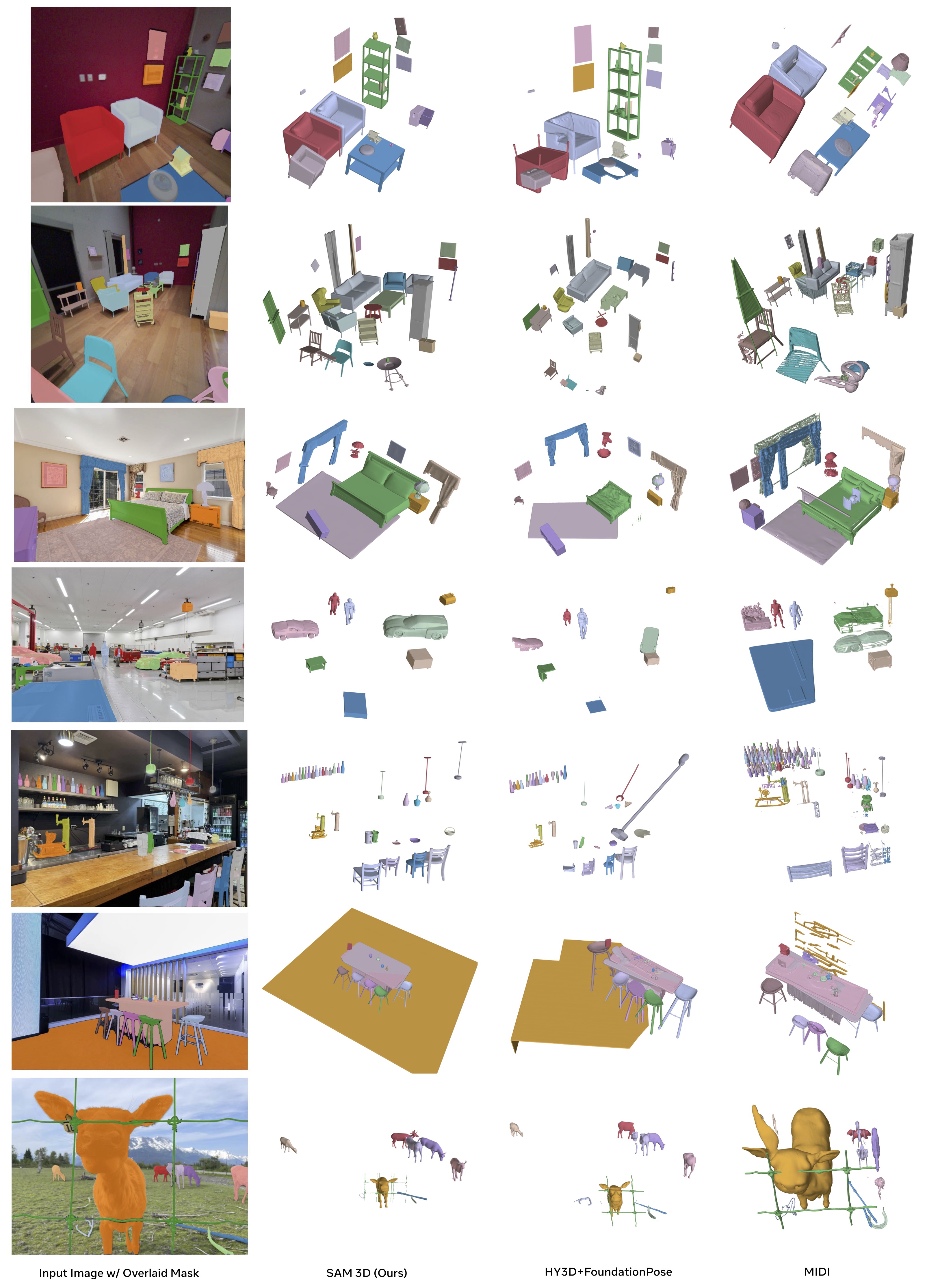}
    \caption{\textbf{Qualitative examples for scene reconstruction.} Showing examples of \method and alternative scene reconstruction methods.}
    \label{fig:layout_visual}
\end{figure*}

\end{document}